\documentclass{article}

\PassOptionsToPackage{numbers, compress}{natbib}

\usepackage[preprint]{neurips_2024}
\usepackage[T5]{fontenc}
\usepackage{graphicx}
\usepackage{multirow}
\usepackage{array} 
\usepackage{booktabs}
\usepackage{multirow}
\usepackage{amsmath}
\usepackage{graphicx}
\usepackage{subcaption}

\usepackage{times}
\usepackage{latexsym}
\usepackage{booktabs}
\usepackage{array}
\usepackage{multirow}
\usepackage{multicol}
\usepackage[numbers]{natbib}

\usepackage[T5]{fontenc}

\usepackage[utf8]{inputenc}
\usepackage{enumitem} 

\usepackage{microtype}

\usepackage{inconsolata}

\usepackage{graphicx}

\usepackage{algorithm}
\usepackage{caption}
\usepackage{algpseudocode}
\usepackage{mathrsfs}
\usepackage{lipsum} 
\usepackage[normalem]{ulem}
\useunder{\uline}{\ul}{}
\usepackage{wrapfig}
\usepackage{float}
\usepackage[colorlinks=true,
            linkcolor=blue,
            citecolor=green,
            filecolor=magenta,
            urlcolor=cyan
            ]{hyperref} 

\usepackage{amsmath}
\usepackage{amsfonts}  
\usepackage{amssymb} 
\usepackage[normalem]{ulem}
\usepackage{dutchcal} 
\usepackage{tipx} 

\let\ul\relax
\usepackage{soul} 
\usepackage{xcolor} 

\usepackage[T5]{fontenc}

\usepackage[utf8]{inputenc}
\usepackage{enumitem} 

\usepackage{microtype}

\usepackage{inconsolata}

\usepackage{graphicx}

\usepackage{lipsum} 
\usepackage[normalem]{ulem}
\useunder{\uline}{\ul}{}
\usepackage{tcolorbox}

\DeclareRobustCommand{\github}{%
  \begingroup\normalfont
  \vspace{0.5em}%
  \raisebox{-0.3em}{%
  \includegraphics[height=1.3em]{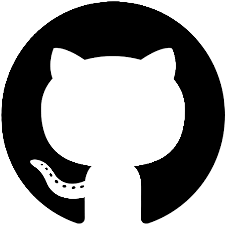}%
  }%
  \kern 0.4em%
  \endgroup
}

\DeclareRobustCommand{\mail}{%
  \begingroup\normalfont
  \vspace{0.0em}%
  \raisebox{0em}{%
  \includegraphics[height=0.8em]{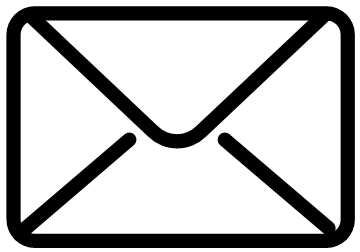}%
  }%
  \kern 0.4em%
  \endgroup
}

\newcommand{\colorhl}[2]{\sethlcolor{#1}\hl{#2}}
\definecolor{custom_light_green}{rgb}{0.85, 0.98, 0.80}

\title{Audio-3DVG: Unified Audio - Point Cloud Fusion for 3D Visual Grounding}
\author{
{\bf Duc Cao-Dinh$^{*1}$ \hspace{0.5cm} Khai Le-Duc$^{*2,3,4}$}
\\ {\bf Anh Dao$^{5}$\hspace{0.5cm} Bach Phan Tat$^{6}$\hspace{0.5cm} Chris Ngo$^{4}$} 
\\ {\bf Duy M. H. Nguyen$^{7,8,9}$ \hspace{0.5cm} Nguyen X. Khanh$^{10}$\hspace{0.5cm}Thanh Nguyen-Tang$^{11}$}\\ 
$^1$Hanyang University, South Korea \hspace{0.2cm}
$^2$University of Toronto, Canada \hspace{0.2cm}\\
$^3$University Health Network, Canada \hspace{0.2cm}
$^4$Knovel Engineering Lab, Singapore \hspace{0.2cm}\\
$^5$Michigan State University, USA \hspace{0.2cm}
$^6$KU Leuven, Belgium\\
$^7$German Research Center for Artificial Intelligence (DFKI), Germany\\
$^8$Max Planck Research School for Intelligent Systems (IMPRS-IS), Germany\\
$^9$University of Stuttgart, Germany \hspace{0.2cm}
$^{10}$UC Berkeley, USA \hspace{0.2cm}
$^{11}$Johns Hopkins University, USA\\
\mail \texttt{duccd@hanyang.ac.kr} \hspace{1 cm} \mail \texttt{duckhai.le@mail.utoronto.ca}\\
\github \href{https://github.com/leduckhai/Audio-3DVG}{\colorhl{custom_light_green}{https://github.com/leduckhai/Audio-3DVG}}
} 

\begin{document}
\maketitle
\begin{abstract}
3D Visual Grounding (3DVG) involves localizing target objects in 3D point clouds based on natural language. While prior work has made strides using textual descriptions, leveraging spoken language-known as Audio-based 3D Visual Grounding-remains underexplored and challenging. Motivated by advances in automatic speech recognition (ASR) and speech representation learning, we propose Audio-3DVG, a simple yet effective framework that integrates audio and spatial information for enhanced grounding. Rather than treating speech as a monolithic input, we decompose the task into two complementary components. First, we introduce \textit{i) Object Mention Detection}, a multi-label classification task that explicitly identifies which objects are referred to in the audio, enabling more structured audio-scene reasoning. Second, we propose an \textit{ii) Audio-Guided Attention} module that models the interactions between target candidates and mentioned objects, enhancing discrimination in cluttered 3D environments. To support benchmarking, we \textit{iii) synthesize audio descriptions for standard 3DVG datasets}, including ScanRefer, Sr3D, and Nr3D. Experimental results demonstrate that Audio-3DVG not only achieves new state-of-the-art performance in audio-based grounding, but also competes with text-based methods, highlight the promise of integrating spoken language into 3D vision tasks. 
\end{abstract}

\def\thefootnote{(*)}\footnotetext{Equal contribution}\def\thefootnote{\arabic{footnote}}

\section{Introduction}

Visual grounding (VG) of referring expressions, the task of identifying visual entities described in natural language, has made significant progress in the 2D computer vision domain~\citep{Liu2019ImprovingRE, Mao2015GenerationAC, Kazemzadeh2014ReferItGameRT, Wang2017LearningTN}. With the rapid advancement of 3D sensing technologies and spatial data representations, this task has naturally extended into the 3D domain, where spatial reasoning becomes increasingly crucial. Unlike 2D images composed of grid-aligned pixels, 3D data, typically represented as point clouds, encode richer geometric and spatial structures. This shift introduces both novel opportunities and unique challenges for accurately grounding language in a three-dimensional space.

In line with this evolution, recent studies have transitioned from grounding objects in 2D images~\citep{Liu2019CLEVRRefDV, Chen2018TOUCHDOWNNL, Chen2020CopsRefAN, Qi2019REVERIERE} to grounding in 3D scenes~\citep{chen2020scanrefer, achlioptas2020referit_3d}, where the goal is to localize objects referenced by natural language within a point cloud (3DVG). Although these advances have yielded strong results, most approaches remain reliant on textual input. This dependence poses a barrier to practical deployment, particularly in hands-busy or eyes-busy environments or for users with motor impairments, highlighting the need for more natural and accessible interaction modalities, such as spoken language, to enable seamless human-robot interaction. 
\begin{figure}[t]
    \centering
    \begin{minipage}{\textwidth}
        \centering
        \begin{subfigure}[b]{0.47\textwidth}
        \includegraphics[width=\linewidth]{figures/main_figures/teaser_1.pdf}
        \caption{Audio-driven visual grounding facilitates essential functions in robotic applications - such as object localization, and autonomous navigation - making it a crucial component for effective human-robot interaction.}
        \label{fig:teaser}
    \end{subfigure}
        \hfill
        \begin{subfigure}[b]{0.45\textwidth}
            \centering
            \includegraphics[width=0.8\linewidth]{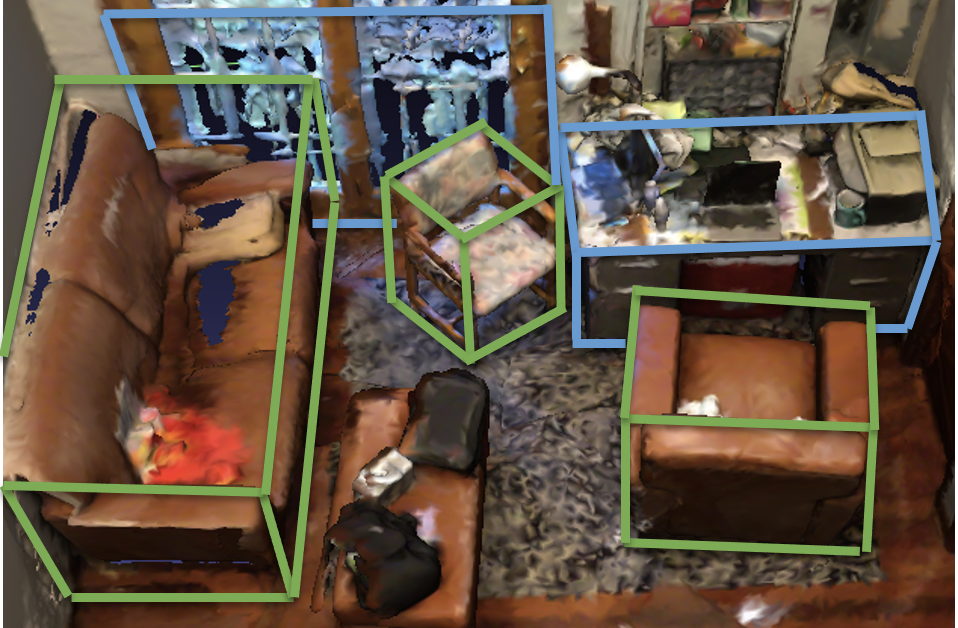} 
            \caption{"The small \textcolor{green!50!black}{chair} has its back to the large \textcolor{blue!70!black}{window}, The chair is next tot the \textcolor{blue!50!black}{desk} in the corner of the room facing the brown \textcolor{blue!50!black}{armchair}". Example sentences referring to objects in 3D scenes. \textcolor{green!50!black}{Green boxes}  indicate objects of the same category as the target, while \textcolor{blue!70!black}{blue boxes} highlight relational objects mentioned in the speech.}
            \label{fig:referring_example}
    \end{subfigure}
    \end{minipage}
\end{figure}

To overcome the limitations of typed text inputs, one practical workaround for enabling voice-based 3D visual grounding is to use an automatic speech recognition (ASR) module to transcribe spoken utterances into text, which can then be fed into existing text-based grounding models, for e.g.,  3DVG-Transformer~\cite{Zhao20213DVGTransformerRM} and InstanceRefer~\cite{Yuan2021InstanceReferCH}. While this allows for speech-driven interaction in practice, it does not fundamentally address the challenge of integrating raw audio signals into the grounding process. Moreover, these ASR-dependent pipelines \textbf{introduce latency and potential transcription errors}, which can hinder real-time interaction and degrade grounding accuracy - particularly in noisy or ambiguous speech scenarios. The decoupling between the speech signal and the 3D visual input also limits the potential for deeper cross-modal alignment between voice and spatial context.

To address this, AP-Refer~\citep{Zhang20243DVG} proposes a fully speech-driven grounding framework that bypasses text transcription entirely. It directly aligns raw point clouds with spoken language to localize referred objects, offering a promising alternative for audio-driven robotic interaction, as illustrated in Figure \ref{fig:teaser}. Despite this innovation, such an approach exhibits two critical limitations. First, it lacks an effective attention mechanism for fusing audio and spatial features, resulting in limited cross-modal understanding. Second, it overlooks relational cues in speech, relying solely on the alignment between the audio input and individual object features. This becomes particularly problematic in cluttered environments where disambiguating between similar objects requires reasoning over spatial and semantic relations.

In this paper, we present a novel framework that directly addresses key limitations in prior audio-based 3D visual grounding methods and significantly narrows the performance gap between audio-based and text-based approaches, all while reducing the latency typically introduced by ASR-dependent pipelines. Our contributions are threefold.

First, to overcome the lack of relational understanding, we observe that target objects are often spatially correlated with other instances of the same category, as well as with additional objects explicitly mentioned in the spoken input (Figure~\ref{fig:referring_example}). To capture this, we introduce an auxiliary task called Object Mention Detection, which identifies the presence of relational objects referenced in the utterance. These relational objects act as spatial anchors, guiding the model to more accurately disambiguate and localize the correct target among multiple candidates.

Second, to enhance cross-modal fusion, we propose an Audio-Guided Attention Module that learns spatial and semantic relationships between candidate objects and their relational counterparts, all conditioned directly on the raw audio signal. Unlike prior methods that rely on intermediate text representations, this attention mechanism allows our model to operate directly on speech, enabling lower-latency inference and reducing error propagation from transcription. It also improves the model’s ability to attend to spatial dependencies, leading to more precise localization.

Third, we contribute a suite of new benchmark datasets for the 3DVG-Audio task, including high-quality synthetic speech derived from existing 3DVG benchmarks, as well as real-world spoken language datasets to evaluate generalization under realistic conditions.

Our experimental results show that the proposed method not only substantially outperforms existing audio-based models, but also achieves performance competitive with state-of-the-art text-based systems while offering faster, end-to-end grounding from voice to prediction.

\section{Related Work}
\subsection{ Multi-modal research based on audio}
Audio, a ubiquitous and natural modality, has long been studied in AI, particularly for tasks like audio classification~\citep{Gemmeke2017AudioSA, Georgescu2022AudiovisualMA, Li2023SelfSupervisedAT} and audio-text fusion~\citep{Lou2022AudioTextRI, Xu2024ADA, Xin2023CooperativeGM}. Recent advances in deep learning have enabled multi-modal models that leverage pre-trained audio representations to solve language and perception tasks. However, in 3D visual grounding, the use of audio remains underexplored. Most existing methods rely on text inputs \cite{Yang2021SAT2S,Zhao20213DVGTransformerRM,Yuan2021InstanceReferCH}, and speech-based approaches often depend on ASR, introducing latency and error propagation. In contrast, our work proposes a novel multi-modal framework that directly aligns raw audio with 3D spatial features, enabling real-time, speech-driven grounding without intermediate transcription. This positions audio as a primary modality for 3D vision-language tasks and opens new directions for human-robot interaction.

While AP-Refer~\citep{Zhang20243DVG} introduced the first audio-point cloud fusion framework for 3D grounding, it lacks explicit mechanisms to model relational object cues or fine-grained spatial dependencies. In comparison, our method incorporates a dedicated Object Mention Detection task and an Audio-Guided Attention Module to capture semantic and geometric relations among objects conditioned on speech, resulting in more precise and robust grounding performance - especially in complex, cluttered scenes.

\subsection{3D Visual Grounding}
3D visual grounding (3DVG) aims to identify objects in 3D scenes based on natural language expressions. Early datasets like ReferIt3D~\citep{achlioptas2020referit_3d} and ScanRefer~\citep{chen2020scanrefer} have spurred significant progress in this domain. Existing methods typically fall into two categories: one-stage approaches that directly fuse language and visual features at the point or patch level to regress bounding boxes~\citep{Luo20223DSPSS3, Kamath2021MDETRM}, and two-stage approaches that follow a detect-then-match pipeline~\citep{Yuan2021InstanceReferCH, Zhao20213DVGTransformerRM}, separating object proposal generation from cross-modal matching for greater interpretability.

While prior works focus predominantly on text-based grounding using either one-stage or two-stage pipelines, our approach introduces a fundamentally different perspective by directly leveraging raw audio input rather than relying on transcribed text. In contrast to existing methods that fuse visual and linguistic features at either the object or point level, our method incorporates an Audio-Guided Attention Module for cross-modal reasoning and an Object Mention Detection task to capture relational cues from speech. This design not only enables end-to-end speech-driven grounding but also enhances relational understanding and reduces inference latency by eliminating the need for intermediate transcription, setting our work apart from the current 3DVG landscape.
\section{Method}
\begin{figure*}
    \centering
    \includegraphics[width=\linewidth]{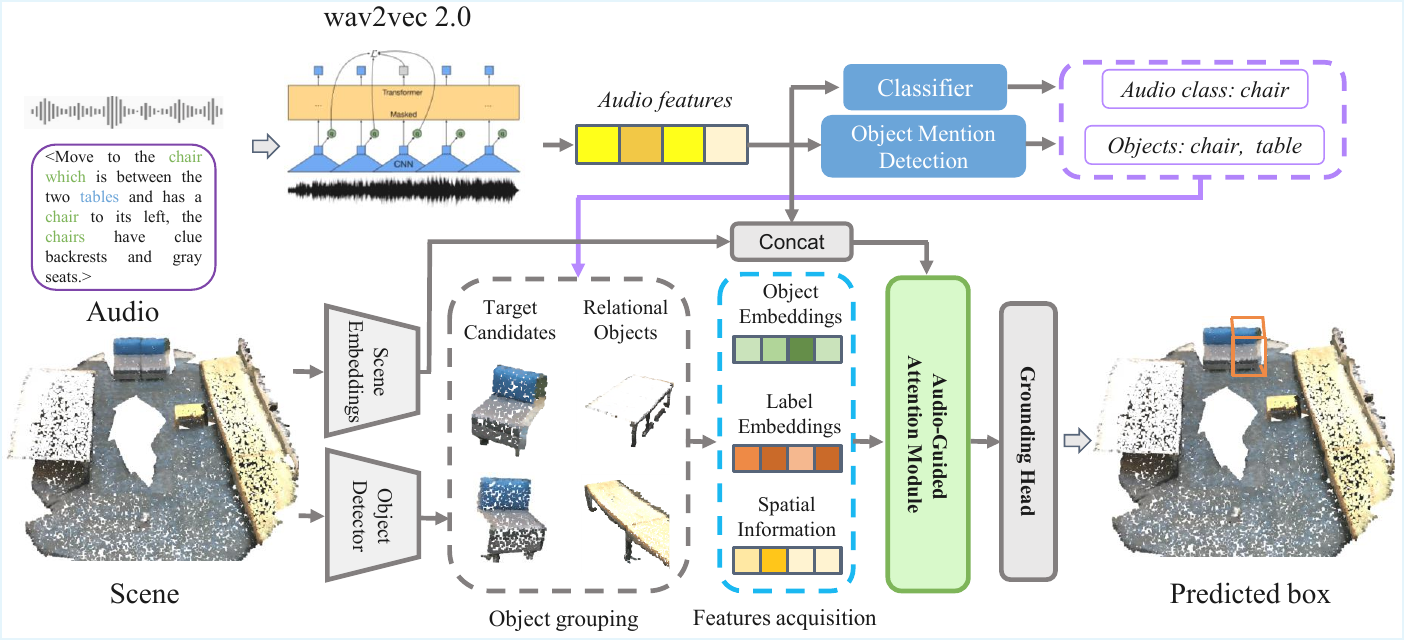}
    \caption{Illustration of the overall pipeline. Our network processes the 3D point cloud using an object detector and encodes the audio input with a wav2vec 2.0 model. The extracted audio and different group object features are then fused via Audio-Guided Attention, which consists of Audio-Guided Self Attention and Audio-Guided Cross Attention modules, followed by a grounding head responsible for target object identification.}
    \label{fig:architecture}
\end{figure*}

Audio-3DVG is a novel framework for audio-based 3D visual grounding that performs target-relation referring to identify the most relevant instance-level object. As illustrated in Figure~\ref{fig:architecture}, the framework leverages point cloud instance segmentation to first extract individual object instances and then construct rich representations for each object within the entire scene. In the upper branch, we utilize the wav2vec 2.0 model~\cite{Baevski2020wav2vec2A} to extract contextual audio representations. These features are then processed by audio classification and Object Mention Detection heads to identify the audio class-used to filter target proposals-and to detect the presence of relational entities within the context. Finally, an Audio-Guided Attention Module is introduced to fuse the multi-modal input representations and guide the selection of the optimal candidate.
\subsection{Instances Generation}
Unlike ScanRefer~\cite{chen2020scanrefer}, which treats all object proposals as potential candidates, our approach follows a recent \textit{detection-then-matching} framework. We first extract all foreground instances from the input point cloud and leverage audio classification to identify a set of likely object candidates. The 3D visual grounding task is then reformulated as an instance-level matching problem. Specifically, given a scence $S$ with point cloud data $P_{scene} \in \mathbb{R}^{Q\times 6}$, we use PointGroup~\citep{Jiang2020PointGroupDP} to detect the object instances present in the scene, producing a set of objects $(o_1,..., o_M )$, each object $o_i$ is represented by a subset of points $P_i \in \mathcal{R}^{K\times6}$ , where each point contains $xyz$ coordinates and $rgb$ color values. In our experiments, we sample $K=1024$ points per object. Each proposal is also associated with a 3D bounding box  $B_T \in \mathcal{R}^{6}$ , encoding the center coordinates and the dimensions of the box.

\subsection{Audio Encoding with Scene Embedding}
Following~\citet{Zhang20243DVG}, Audio-3DVG employs an ASR pre-trained wav2vec 2.0 base model~\citep{Baevski2020wav2vec2A} for audio feature extraction (see Appendix Section \ref{sec:ASR}). wav2vec 2.0 is an unsupervised speech representation learning framework that has shown strong performance across a wide range of speech-related downstream tasks (see Appendix Section \ref{sec:SSL_speech} for a detailed discussion of our rationale). Given an input audio signal $\mathcal{A}_T$, wav2vec 2.0 produces a feature representation $F_{Audio}\in \mathbb{R}^{C \times L_a}$ , where $L_a$ is the sequence length and $C$ is the feature dimensionality of the high-dimensional latent space. To further encode temporal dependencies and contextual information, $F_{Audio}$ is passed through bidirectional GRU layers, resulting in a fixed-length 768-dimensional vector ($a$) used for downstream optimization. 

To incorporate scene-level geometric context, we also embedd the raw scene point cloud using a sparse convolutional neural network. Specifically, we employ the Minkowski Engine \cite{choy20194d}, a highly efficient library for processing sparse tensor data, to extract spatial features from the 3D scene. The point cloud is voxelized and passed through a series of sparse convolutional layers to produce a global scene representation. This results in a compact 512-dimensional feature vector, which is concatenated with the audio features to capture the overall structural layout of the environment.
\subsection{Audio Classification}
Leveraging the contextualized audio representation, we design a classifier to predict the target object referenced in the spoken utterance. The classifier is implemented as a lightweight multilayer perceptron (MLP) followed by a softmax layer to produce class probabilities. The MLP includes $N_{cls}$ output heads, corresponding to the total number of unique object classes defined in the dataset.

\subsection{Object Mention Detection}
Most prior works~\citep{chen2020scanrefer, achlioptas2020referit_3d, Yuan2021InstanceReferCH} focus solely on analyzing candidate object proposals, often neglecting the presence of relational objects referenced in the input. This limitation can lead to ambiguity when identifying the correct target in scenes with dense object instances. To address this issue, we propose a novel auxiliary task called Object Mention Detection, which aims to identify relational objects mentioned in the audio.
Concretely, we employ a lightweight multilayer perceptron (MLP) with 
$N_{cls}$ binary classification heads, corresponding to maximum $N_{cls}$ object classes may appear in the scene, Each head predicts the probability that its corresponding object class is mentioned in the spoken utterance. During inference, objects with predicted probabilities exceeding a predefined threshold are classified as relational objects.
\subsection{Object Grouping}
Before passing instances to subsequent modules, Audio-3DVG leverages the predicted instance set from earlier stages to filter candidate objects and identify relevant relational references. For example, given an audio description such as \textit{‘The chair is between two tables and has a chair to its left, the chairs has clue backrest and a gray seat’}, we begin with all object instances extracted from the original point cloud $(o_1,..., o_M )$. . From this set, we retain only those instances classified as the target category, \textit{‘chair’}, along with the related category, \textit{‘table’}. These filtered sets correspond to the \textit{target candidates} point set ( $P^N_{i=0} \subset P_{Target}$, where each $P_i\in \mathcal{R}^{1024\times6}$), and the \textit{relational objects} point set ( $P^N_{j=0}\subset P_{Relational}$, where each $P_j\in\mathcal{R}^{1024\times6}$) in our pipeline.
\subsection{Object Feature Acquisition}
Unlike \citet{chen2022vil3dref}, who explicitly decouple semantic and spatial information in object representations, we adopt a unified embedding that integrates multiple feature modalities through concatenation, including:

\textbf{Object Embedding:} For each object $o_i$ in the set of target candidates and relational objects is represented as $P_i\in\mathcal{R}^{K\times(3+F)}$, where $K$ denotes the number of points, 3 corresponds to the spatial coordinates ($x, y, z$) associated with each point, and $F$ includes additional point-wise attributes such as RGB color values in our case. we first normalize the coordinates of its point cloud into a unit ball. We then use PointNet++~\citep{qi2017pointnet++} - a widely adopted framework for 3D semantic segmentation and object detection to extract object-level features, resulting in $o^{obj}_i\in\mathcal{R}^{1\times1024}$.

\textbf{Label Embedding:} To enhance the target classifier’s awareness of candidate categories, ReferIt3D~\cite{achlioptas2020referit_3d} incorporates an auxiliary classification task within a joint optimization framework. Although this approach improves category understanding, it also adds to the overall learning complexity. In our network, we incorporate object labels as part of the object representation by embedding them using word embedding model. Specifically, for each instance in the set of target candidates and relational objects, we encode its class label using a pre-trained GloVe~\citep{Pennington2014GloVeGV}, resulting $o^{label}_i\in\mathcal{R}^{1\times300}$.

\textbf{Spatial Information:} To represent the absolute position of each instance $O_i$ with corresponding representation $P_i$, we compute the object center $o^{center}_i = [c_x, c_y, c_z] \in \mathcal{R}^{3}$ and the object size $o^{size}_i = [z_x, z_y, z_z]\in \mathcal{R}^{3}$ These are derived from the object points $P_i$
where the center is calculated as the mean of $P_i$, and the size corresponds to the spatial extent of $P_i$.

All of these features are concatenated into a single representation:
\begin{equation}
    o^{rep}_i = [o^{obj}_i, o^{label}_i, o^{center}_i, o^{size}_i]
\end{equation}
\subsection{Audio-Guided Attention Module}
\begin{figure}[t]
    \centering
    \begin{minipage}{\textwidth}
        \centering
        \begin{subfigure}[b]{0.43\textwidth}
        \centering
        \includegraphics[width=\linewidth]{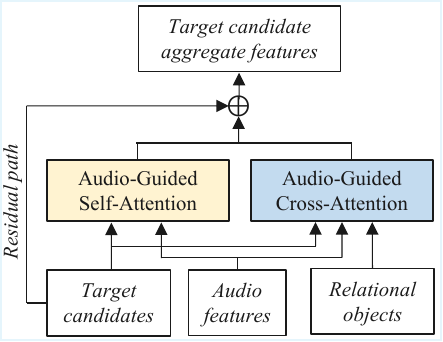} 
        \caption{Audio-Guided Attention overview.}
        \label{fig:attention_overview}
    \end{subfigure}
        \hfill
        \begin{subfigure}[b]{0.53\textwidth}
            \centering
            \includegraphics[width=\linewidth]{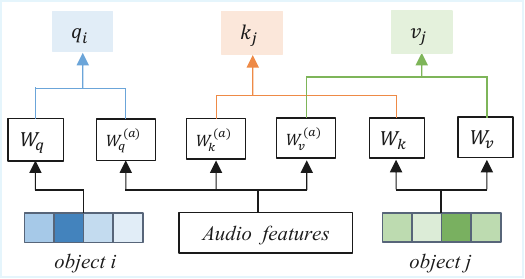} 
            \caption{Given two objects, this figure illustrates the mechanism for computing the key, query, and value vectors used to estimate the attention score between them.}
            \label{fig:attention_detail}
        \end{subfigure}
    \end{minipage}
    \caption{An overview of the proposed Audio-Guided Attention Module, comprising the Audio-Guided self-Attention and Audio-Guided cross-Attention submodules designed to model contextual and relational cues from speech.}
    \label{fig:example_1}
\end{figure}
In both text and audio descriptions, the target object is often identified through references to relational objects (e.g., “the chair is in front of the door, opposite to the coffee table”) or through spatial comparisons with objects of the same category (e.g., “of the two brown wooden doors, choose the door on the left when facing them”). Building on this observation, after obtaining feature representations for each target candidates and relational objects, We design an attention module comprising two components: \textbf{Audio-Guided Self-Attention}, which helps distinguish the target object from other instances within the same category, and \textbf{Audio-Guided Cross-Attention}, which captures spatial relationships between target candidates and relational objects referenced in the audio, as illustrated in Figure~\ref{fig:attention_overview}.
Specifically, as shown the the Figure~\ref{fig:attention_detail}, given audio feature $a\in\mathcal{R}^{d_a}$, and a pair object $o_i\in\mathcal{R}^{1\times d}$ and $o_j\in\mathcal{R}^{1\times d}$, each attention module first compute the embedding by projecting object, each modulated by audio feature in to query, key, and value spaces $ \mathbf{q}_i = W_qo_i + W^{(a)}_qa$,\ $\mathbf{k}_j = W_ko_j + W^{(a)}_ka$,\,$ \mathbf{v}_j = W_vo_j + W^{(a)}_va$.
Then the standard scaled dot-product attention value is calculated as:
\begin{equation}
\text{attention}_{ij} = \frac{\mathbf{q}_i^\top \mathbf{k}_j}{\sqrt{d}} 
\quad \Rightarrow \quad 
\alpha_{ij} = \mathrm{softmax}_j(\text{attention}_{ij})
\end{equation}
Then:\ $
\mathbf{o}'_i = \sum_{j=1}^{N} \alpha_{ij} \mathbf{v}_j$. 
Each output $\mathbf{o}'_i$ is audio-modulated feature representing the object with the context from other object. In the implementation, we use stack \textbf{multi-head attention} and concatenate the outputs:
\begin{equation}
\mathrm{MultiHead}(O, a) = \mathrm{Concat}(\text{head}_1, \ldots, \text{head}_h) W_o
\end{equation}
With $W_o$ is learnable weights, the audio-guided attention scores are computed between object pairs in the Audio-Guided Self-Attention module, resulting in $\mathbf{O}' = { \mathbf{o}'_1, \ldots, \mathbf{o}'_N }$. In contrast, the Audio-Guided Cross-Attention module computes attention scores between each target candidate and all relational objects mentioned in the audio, producing $\mathbf{O}'' = { \mathbf{o}''_1, \ldots, \mathbf{o}''_N }$. Finally, the aggregated feature representation for each target candidate is obtained by summarizing the features from $\mathbf{O}$, $\mathbf{O}'$, and $\mathbf{O}''$.
\subsection{Grounding Head}
Finally, we employ a classifier to identify the target object referenced in the speech. This classifier consists of a multilayer perceptron (MLP) followed by a softmax layer, which predicts the most likely target among the $N$ candidate objects.
\subsection{Loss functions}
We employ multiple loss functions to train Audio-3DVG across its various tasks, including an audio classification loss $\mathcal{L}^{cls}_{audio}$, a multi-label classification loss for the Object Mention Detection task $\mathcal{L}^{OMD}_{audio}$, and an object classification loss for the grounding task $\mathcal{L}^{cls}_{object}$. Therefore, the overall training objective is as follows: 
\begin{equation}
    \mathcal{L} = \lambda_a \mathcal{L}^{cls}_{audio} + \lambda_b \mathcal{L}^{OMD}_{audio} + \lambda_c  \mathcal{L}^{cls}_{object}
\end{equation}
where $\lambda_a$, $\lambda_b$, and $\lambda_c$ are three hyper-parameters to balance the losses.
\section{Datasets}
\textbf{ScanRefer~\cite{chen2020scanrefer}:} The dataset contains 51,583 human-written sentences annotated for 800 scenes in ScanNet dataset~\cite{Dai2017ScanNetR3}. Following the official split, we use 36,665 samples for training and 9,508 for validation. Based on whether the target object belongs to a unique category within the scene, the dataset is further divided into two subsets: "unique", where the target class appears only once, and "multiple", where it appears more than once.

\textbf{Nr3D~\cite{achlioptas2020referit_3d}:} The dataset comprises 37,842 human-written sentences that refer to annotated objects in 3D indoor scenes from the ScanNet dataset~\cite{Dai2017ScanNetR3}. It includes 641 scenes, with 511 used for training and 130 for validation, covering a total of 76 target object classes. Each sentence is crafted to refer to an object surrounded by multiple same-class distractors. For evaluation, the sentences are divided into "easy" and "hard" subsets: in the easy subset, the target object has only one same-class distractor, whereas in the hard subset, multiple distractors are present. Additionally, the dataset is categorized into "view-dependent" and "view-independent" subsets, based on whether grounding the referred object requires a specific viewpoint. 

\textbf{Sr3D~\cite{achlioptas2020referit_3d}:} This dataset is constructed using sentence templates to automatically generate referring expressions. These sentences rely solely on spatial relationships to distinguish between objects of the same class. It contains 1,018 training scenes and 255 validation scenes from ScanNet dataset~\cite{Dai2017ScanNetR3}, with a total of 83,570 sentences. For evaluation, it can be partitioned in the same manner as the Nr3D dataset.

To address the data scarcity issue in the Audio-3D visual grounding task, we efficiently convert ScanRefer’s natural language descriptions into audio using Spark-TTS~\cite{Wang2025SparkTTSAE}, an advanced and flexible text-to-speech system that leverages large language models (LLMs) to generate highly accurate and natural-sounding speech. The detailed analysis and configuration of the generated data are presented in the appendix.
\section{Experiments}
\subsection{Experimental Setup}
\textbf{Evaluation Metrics:} We evaluate models under two evaluation settings. One uses ground-truth object proposals, which is the default setting in the Nr3D and Sr3D datasets. The metric is the accuracy of selecting the target bounding box among the proposals. The other setting does not provide ground-truth object proposals and requires the model to regress a 3D bounding box, which is the default setting for the ScanRefer dataset. The evaluation metrics are acc@0.25 and acc@0.5, which is the percentage of correctly predicted bounding boxes whose IoU is larger than 0.25 or 0.5 with the ground-truth.

\textbf{Implementation details.} We adopt the official pre-trained PointGroup~\citep{Jiang2020PointGroupDP} as the backbone for instance segmentation. For audio encoding, we utilize a BiGRU to extract word-level features with a channel dimension of 768. All employed MLPs use hidden layers configured as [521,64], followed by Batch Normalization and ReLU activation. We use 8 attention heads, each producing features with a dimensionality of 128. The network is trained for 30 epochs using the Adam optimizer with a batch size of 32. The learning rate is initialized at 0.0005 and decayed by a factor of 0.9 every 5 epochs. All experiments are implemented in PyTorch and run on a single NVIDIA RTX 3090 GPU.

\subsection{Experimental Results}
We begin by presenting the performance results for the audio classification and object-mention detection tasks across three datasets. The Audio Classification task achieves a high accuracy of 96\%, demonstrating the reliability of our speech understanding component. For the Object Mention Detection task, we report the average precision, recall, and F1 score as provided in Table~\ref{tab:table_4}. The overall quantitative performance is constrained by the scarcity of a few rare object classes. However, for common categories, the F1-score reaches approximately 97\%, ensuring strong reliability for downstream tasks.
For the 3D visual grounding performance, we present the comparative results on the ScanRefer dataset using detected objects from PointGroup~\cite{Jiang2020PointGroupDP}. Given the same audio input, our model demonstrates a substantial performance improvement over AP-Refer, highlighting the effectiveness of our approach. Furthermore, our method achieves competitive results compared to text-based methods. However, it is important to note that this comparison is not entirely fair, as text inputs provide richer and error-free linguistic information, whereas our audio-based approach is subject to potential inaccuracies introduced during text-to-speech conversion. To enable a fairer comparison, we convert the audio inputs back to text by Whisper~\cite{Radford2022RobustSR} and evaluate the performance of recent state-of-the-art methods, as reported in Table~\ref{tab:table_1}. 
Next, we evaluate our method on the Nr3D and Sr3D datasets, as presented in Table~\ref{tab:table_2}. Our approach demonstrates clear improvements over the baseline~\citep{achlioptas2020referit_3d}, reinforcing the potential of voice as a viable and effective alternative to text for 3D visual grounding tasks.
Lastly, we evaluate our network on real-world datasets, establishing a new benchmark for assessing audio-based 3D visual grounding in practical applications, as presented in Table~\ref{tab:table_7}.

\begin{table*}[ht]
\centering
    \tiny
    \setlength{\tabcolsep}{2pt} 
    \caption{Comparison of grounding accuracy across different methods under different inputs.}
    \label{tab:table_1}
    \resizebox{0.8\textwidth}{!}{%
    \begin{tabular}{lcccccccc}
        \toprule
        Method & Input & \multicolumn{2}{c}{Unique} & \multicolumn{2}{c}{Multiple} & \multicolumn{2}{c}{Overall} \\
        & & acc@0.25 & acc@0.5 & acc@0.25 & acc@0.5 & acc@0.25 & acc@0.5 \\
        \midrule
        ScanRefer        & Text        & 65.00 & 43.31 & 30.63 & 19.75 & 37.30 & 24.32 \\
        TGNN            & Text        & 68.61 & 56.80 & 29.84 & 23.18 & 37.37 & 29.70 \\
        Non-SAT         & Text        & 68.48 & 47.38 & 31.81 & 21.34 & 38.92 & 26.40 \\
        {SAT}    & Text        & 73.21 & 50.83 & 37.64 & 25.16 & 44.54 & 30.14 \\
        3DVG-Trans      & Text        & 77.16 & 58.47 & 38.38 & 28.70 & 45.90 & 34.47 \\
        InstanceRefer   & Text        & 78.37 & 66.88 & 27.90 & 21.83 & 37.69 & 30.57 \\
        \midrule
        {3DVG-Trans}     & Audio2Text & 74.92 & 56.67 & 35.43 & 26.92 & 43.23 & 33.87 \\
        {InstanceRefer}  & Audio2Text & 73.28 & 64.20 & 29.12 & 22.98 & 38.46 & 30.90 \\
        AP-Refer        & Audio       & 48.62 & 29.59 & 16.94 & 9.96  & 23.09 & 13.77 \\
        \textbf{Ours}   & Audio       & \textbf{78.26} & \textbf{67.35} & \textbf{30.82} & \textbf{24.46} & \textbf{40.02} & \textbf{32.78} \\
        \bottomrule
    \end{tabular}
    }
\end{table*}

\begin{table*}[ht]
    \centering
    \tiny
    \setlength{\tabcolsep}{2pt} 
    \caption{Comparison of grounding accuracy across different methods under different inputs.}
    \label{tab:table_2}
    \centering
    \resizebox{0.8\textwidth}{!}{%
        \setlength{\tabcolsep}{2pt} 
        \begin{tabular}{l  c c c c c c c c c c c}
        \toprule
        \multirow{2}{*}{Method} & \multirow{2}{*}{Input} & \multicolumn{5}{c}{Nr3D} & \multicolumn{5}{c}{Sr3D} \\
         &  & Overall & Easy & Hard & \shortstack{View\\Dep} & \shortstack{View\\Indep} & Overall & Easy & Hard & \shortstack{View\\Dep} & \shortstack{View\\Indep} \\
        \midrule
        ReferIt3D  & Text & 35.6 & 43.6 & 27.9 & 32.5 & 37.1 & 40.8 & 44.7 & 31.5 & 39.2 & 40.8 \\
        ScanRefer & Text& 34.2 & 41.0 & 23.5 & 29.9 & 35.4 & - & - & - & - & - \\
        \midrule
        \textbf{Ours}  & \textbf{Audio} & \textbf{37.4} & \textbf{45.2} & \textbf{30.9} & \textbf{34.1} & \textbf{40.7} & \textbf{48.3} & \textbf{51.3} & \textbf{40.9} & \textbf{45.1} & \textbf{48.6} \\
        \bottomrule
        \end{tabular}%
        }
\end{table*}

\begin{table*}[!hbt]
    \centering
    \begin{minipage}[t]{0.5\linewidth}
        \centering
        \tiny
        \setlength{\tabcolsep}{2pt} 
        \caption{Performance on the Object Mention Detection task is evaluated separately for ScanRefer, Nr3D, and Sr3D dataset using precision, recall, and F1 score metrics.}
        \label{tab:table_3}
        \centering
        \resizebox{0.65\textwidth}{!}{%
            \setlength{\tabcolsep}{2pt} 
            \begin{tabular}{l  c c c }
            \toprule
            & ScanRefer & Nr3D & Sr3D \\
            \midrule
            Precision & 0.72 & 0.75 & 0.77\\
            Recall & 0.71 & 0.77 & 0.77\\
            F1 & 0.71 & 0.76 & 0.77\\
            \bottomrule
            \end{tabular}%
            }
    \end{minipage}%
    \hfill
    \begin{minipage}[t]{0.45\linewidth}
        \centering
        \tiny
        \setlength{\tabcolsep}{2pt} 
        \caption{Response time comparison across different pipelines for 3D visual grounding. 
        }
        \label{tab:table_4}
        \centering
        \resizebox{0.9\textwidth}{!}{%
            \setlength{\tabcolsep}{2pt} 
            \begin{tabular}{l  c c c }
            \toprule
            & Text & Voice to text & Voice \\
            \midrule
            Input text &  1300 ms & x & x \\
            Input voice  &  x & 750 ms & 750 ms\\
            Speech to text  & x & 250 ms & x\\
            Model running & 200 ms & 200 ms & 300 ms \\
            \midrule
            Total  & 1500 ms & 1200 ms & 1050 ms\\
            \bottomrule
            \end{tabular}%
            }

    \end{minipage}
\end{table*}
\begin{figure}
    \centering
    \caption{Qualitative results from our network and ScanRefer on the ScanNet dataset (first example), and on the ReferIt3D using Sr3D (second example) and Nr3D (last example). Zoom for clear view.}
    \includegraphics[width=0.7\linewidth]{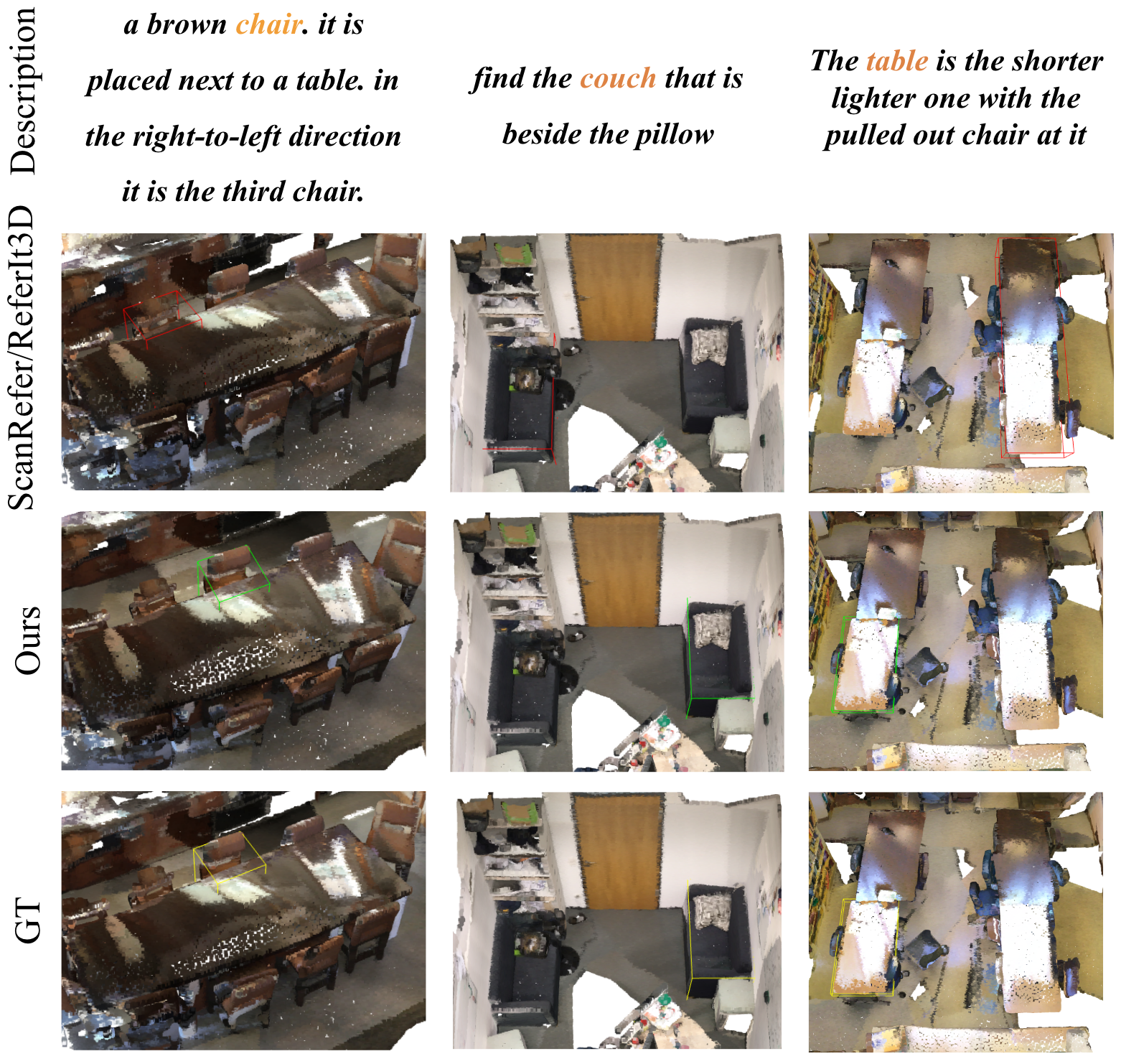}
    \label{fig:result}
\end{figure}
\section{Ablation Study}
\subsection{Result with different audio generation methods}
We further evaluate the performance of Audio-3DVG on the ScanRefer dataset using different text-to-speech (TTS) methods. Following the AP-Refer setup, we replace Spark-TTS with Matcha-TTS~\cite{Mehta2023MatchaTTSAF} to generate audio inputs for this experiment. The results, shown in Table \ref{tab:table_5}, indicate that although training the model with Matcha-TTS leads to slightly lower performance compared to Spark-TTS, it still surpasses the performance of AP-Refer~\cite{Zhang20243DVG} as reported in Table \ref{tab:table_1}.
\begin{table}[ht]
\captionsetup{skip=10pt}
\caption{Performance of Audio-3DVG on ScanRefer dataset with different TTS methods}
\centering
\resizebox{0.6\textwidth}{!}{%
\setlength{\tabcolsep}{3pt}
\begin{tabular}{lccccccc}
\toprule
TTS method   & \multicolumn{2}{c}{Unique} & \multicolumn{2}{c}{Multiple} & \multicolumn{2}{c}{Overall} \\
  & acc@0.25 & acc@0.5 & acc@0.25 & acc@0.5 & acc@0.25 & acc@0.5 \\
\midrule
Matcha-TTS& 76.80 & 64.02 & 30.08 & 23.12 & 39.14 & 31.05 \\
Spark-TTS& \textbf{78.26} & \textbf{67.35} & \textbf{30.82} & \textbf{24.46} & \textbf{40.02} & \textbf{32.78}\\
\bottomrule
\end{tabular}%
}
\label{tab:table_5}
\end{table}
\subsection{Impact of Audio-Guided Attention}
We evaluate the Audio-Guided Attention module on the ScanRefer dataset by replacing it with a baseline MLP that concatenates audio and object features without structured interaction. We also integrate our module into InstanceRefer. As shown in Table~\ref{tab:table_6}, our design highlights its effectiveness.
\subsection{Response time measurement}
To demonstrate the advantages of using voice in 3D visual grounding tasks, we conduct a response time analysis across multiple input modalities and system configurations. First, we measure the response time of a traditional text-based pipeline, which includes both user input time and the inference time of the grounding model. Second, we evaluate a two-stage pipeline where spoken input is first transcribed using a speech-to-text (STT) system, followed by inference using the same text-based model. Finally, we assess our proposed end-to-end audio-driven grounding framework, which directly takes voice input and predicts the target object without intermediate transcription. All experiments are conducted on a sample dataset comprising 100 records, each containing an average of 20 words per query. For the speech-to-text (ASR) component, we use wav2vec 2.0, while Whisper is used as the STT baseline. We leverage InstanceRefer as a typical text-based solution. The results, summarized in Table~\ref{tab:table_4}, highlight the efficiency, responsiveness, and user-friendliness of using spoken language as a natural interface for 3D visual grounding.

\begin{table}[ht]
\captionsetup{skip=10pt}
\caption{Ablation study on the ScanRefer dataset demonstrating the effectiveness of the proposed Audio-Guided Attention (AGA) module.}
\centering
\resizebox{0.75\textwidth}{!}{%
\begin{tabular}{lccccccc}
\toprule
Method   & \multicolumn{2}{c}{Unique} & \multicolumn{2}{c}{Multiple} & \multicolumn{2}{c}{Overall} \\
  & acc@0.25 & acc@0.5 & acc@0.25 & acc@0.5 & acc@0.25 & acc@0.5 \\
\midrule
Ours+MLP& 70.12 & 58.41 & 27.08 & 21.20 & 35.42 & 28.42 \\
Ours+AGA & \textbf{78.26} & \textbf{67.35} & \textbf{30.82} & \textbf{24.46} & \textbf{40.02} & \textbf{32.78}\\
\midrule
InstanceRefer&78.37 & 66.88 & 27.90 & 21.83 & 37.69 & 30.57 \\
InstanceRefer+AGA&\textbf{78.83} & \textbf{67.92} & \textbf{34.18} & \textbf{26.04} & \textbf{42.84} & \textbf{34.16} \\
\bottomrule
\end{tabular}%
}
\label{tab:table_6}
\end{table}

\begin{table}[ht]
\captionsetup{skip=10pt}
\caption{Overall performance on real-world datasets. Each dataset contains approximately 150–200 samples. We plan to release these datasets upon paper acceptance.}
\centering
\tiny
\resizebox{0.5\textwidth}{!}{%
\begin{tabular}{lccc}
\toprule
{\tiny Metrics/Datasets}   & {\tiny ScanRefer} & {\tiny Nr3D} & {\tiny Sr3D} \\
\midrule
{\tiny acc@0.25} & {\tiny 36.72} &x & x \\
{\tiny acc@0.50} & {\tiny 30.08} & x & x \\
Overall & x & {\tiny 35.2} & {\tiny 46.8} \\
\bottomrule
\end{tabular}%
}
\label{tab:table_7}
\end{table}
\section{Limitations}
Despite its effectiveness, audio-based 3D visual grounding still faces notable limitations. First, the Object Mention Detection module struggles with rare object classes due to class imbalance in the dataset (Table~\ref{tab:table_3}), which limits its ability to generalize. This could be mitigated by developing more balanced and diverse datasets. Additionally, the performance of our approach is closely tied to the quality of 3D object segmentation; using more robust segmentation models could significantly enhance grounding accuracy.

Moreover, while our method reduces latency by avoiding ASR, its performance still falls short of state-of-the-art text-based models that benefit from mature language understanding. However, this gap can be narrowed by integrating our audio-driven framework with existing text-based methods. For example, audio-guided relational cues could complement proposal generation or refinement in text-based pipelines, enabling a more robust and multimodal grounding system.
\section{Conclusion}
In summary, this work introduces a novel approach that leverages audio for the 3D visual grounding task. Our contributions include a method for detecting target candidates and relational objects, an effective feature formulation strategy, and a robust attention module for identifying targets within dense object scenes. Additionally, we provide a synthetic audio dataset to support future research in this area.
Our results demonstrate the effectiveness of using audio for 3D vision tasks and highlight its potential as a promising direction for future exploration.

\section{Acknowledgement}

Most of the ASR theory in this work were borrowed from lectures by Prof. Hermann Ney, PD Ralf Schlüter, and PhD Albert Zeyer, as well as from PhD dissertations and especially master thesis by Minh Nghia Phan at RWTH Aachen University.

\newpage

\clearpage 

\appendix

\tableofcontents
\newpage

\section{Details of Speech Dataset}
\subsection{Real-world Speech Data}
\begin{table}[h]
  \centering
  \begin{tabular}{l|cccc}
 & \textbf{\#Speakers} & \textbf{\#Nationalities} & \textbf{\#Samples} & \textbf{\#Genders} \\ \hline
ScanRefer & 5 & 3 & 150 & 2 \\
Nr3D & 6 & 2 & 180 & 2 \\
Sr3D & 7 & 3 & 219 & 2 \\ \hline
\textbf{All} & 9 & 3 & 549 & 2 \\ \hline
\end{tabular}
  \caption{Data statistics of our collected real-world speech dataset. This real-world dataset is only used as \textbf{\textit{test set}}. Recordings were conducted by a group of nine speakers in total, including six Vietnamese (3 females and 3 males), two Belgian (2 males), and one Chinese (female). The age range of the participants spanned from 21 to 31 years.}
  \label{tab:data_stats_realworld}
\end{table}
For this project, a diverse dataset of 549 samples (sentences) was carefully gathered, as shown in Table \ref{tab:data_stats_realworld}. This corpus comprised 150 samples sourced from the ScanRefer text dataset, 180 samples  from the Nr3D dataset and an additional 219 samples from the Sr3D dataset. 

To ensure a rich variety of speech patterns, recordings were conducted by a group of nine speakers, including six Vietnamese (3 females and 3 males), two Belgian (2 males), and one Chinese (female). The age range of the participants spanned from 21 to 31 years. While the selection of speakers was primarily based on convenience, conscious efforts were made to incorporate diversity in accents, gender, and age, thereby enhancing the representativeness of the dataset. Throughout the recording process, we tried to control the noises (e.g. background, echo) and ensure pronunciation clarity, contributing to the overall quality of the collected data.

\begin{figure}[h]
    \centering
    \includegraphics[width=0.8\linewidth]{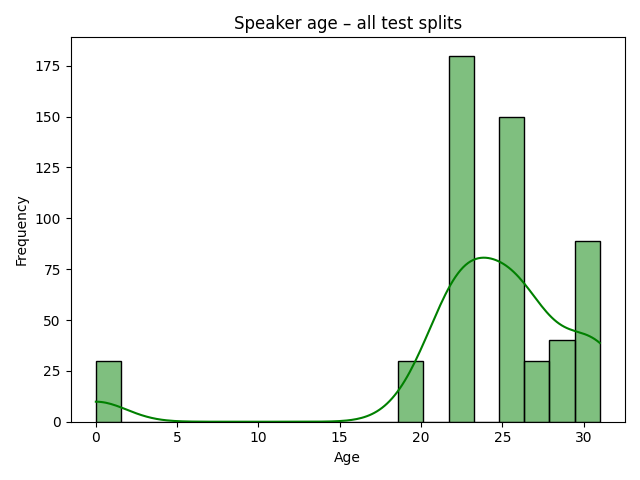}
    \caption{Age distribution of the entire real-world dataset, including ScanRefer, Nr3D and Sr3D.}
    \label{fig:age_all}
\end{figure}

\begin{figure}[h]
    \centering
    \includegraphics[width=0.8\linewidth]{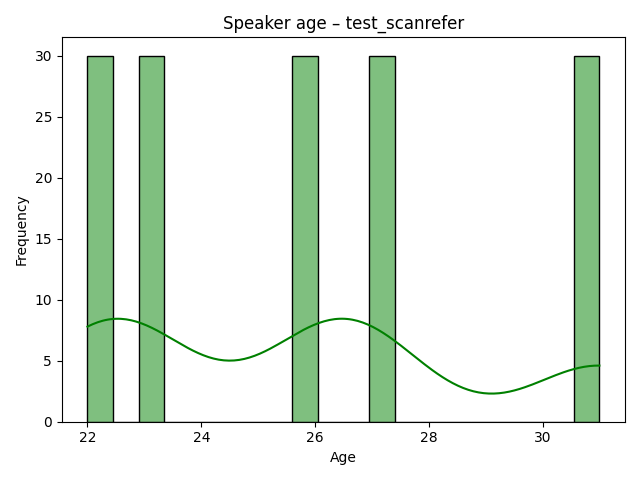}
    \caption{Age distribution of the real-world ScanRefer set.}
    \label{fig:age_test_scanrefer}
\end{figure}

\begin{figure}[h]
    \centering
    \includegraphics[width=0.8\linewidth]{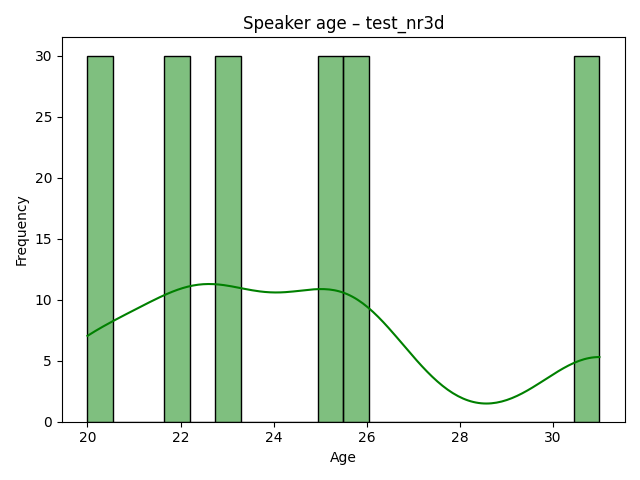}
    \caption{Age distribution of the real-world Nr3D set.}
    \label{fig:age_test_nr3d}
\end{figure}

\begin{figure}[h]
    \centering
    \includegraphics[width=0.8\linewidth]{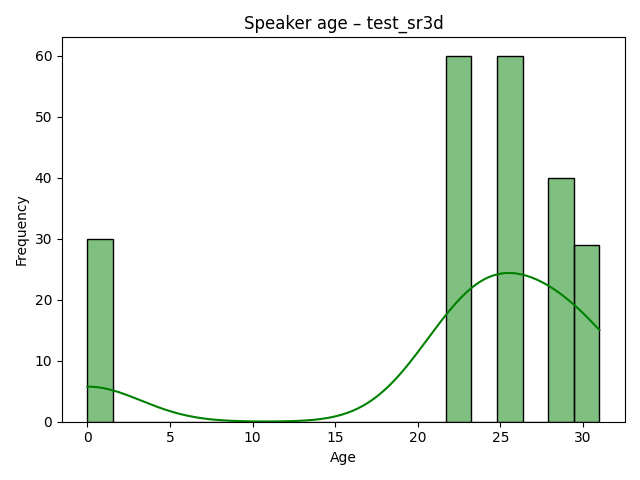}
    \caption{Age distribution of the real-world Sr3D set.}
    \label{fig:age_test_sr3d}
\end{figure}

\onecolumn
\textbf{Age distribution}: Table \ref{fig:age_all}, \ref{fig:age_test_scanrefer}, \ref{fig:age_test_nr3d} and \ref{fig:age_test_sr3d} show the plots of age distribution in the entire real-world dataset, ScanRefer set, Nr3D set and Sr3D set respectively.

\begin{figure}[h]
    \centering
    \includegraphics[width=0.8\linewidth]{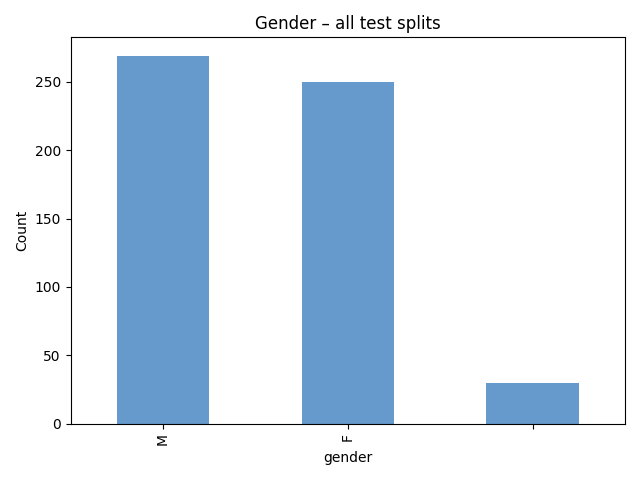}
    \caption{Gender distribution of the entire real-world dataset, including ScanRefer, Nr3D and Sr3D.}
    \label{fig:gender_all}
\end{figure}

\begin{figure}[h]
    \centering
    \includegraphics[width=0.8\linewidth]{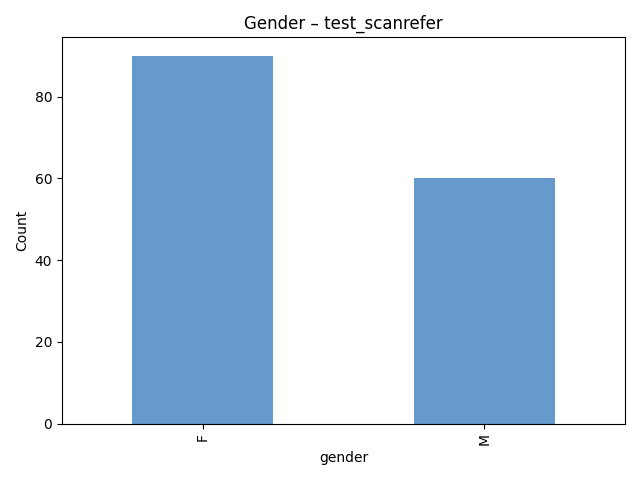}
    \caption{Gender distribution of the real-world ScanRefer set.}
    \label{fig:gender_test_scanrefer}
\end{figure}

\begin{figure}[h]
    \centering
    \includegraphics[width=0.8\linewidth]{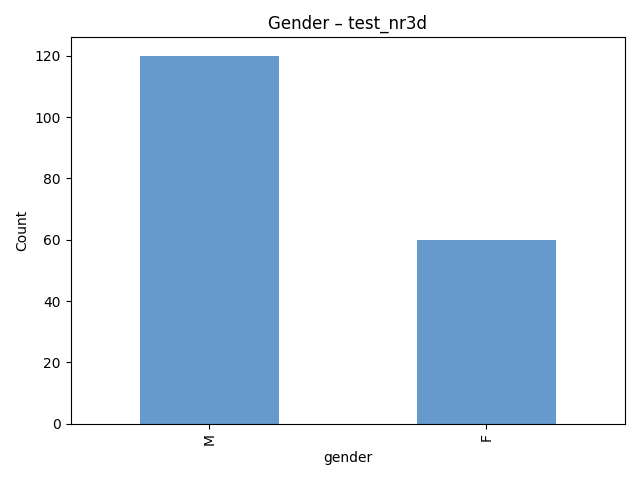}
    \caption{Gender distribution of the real-world Nr3D set.}
    \label{fig:gender_test_nr3d}
\end{figure}

\begin{figure}[h]
    \centering
    \includegraphics[width=0.8\linewidth]{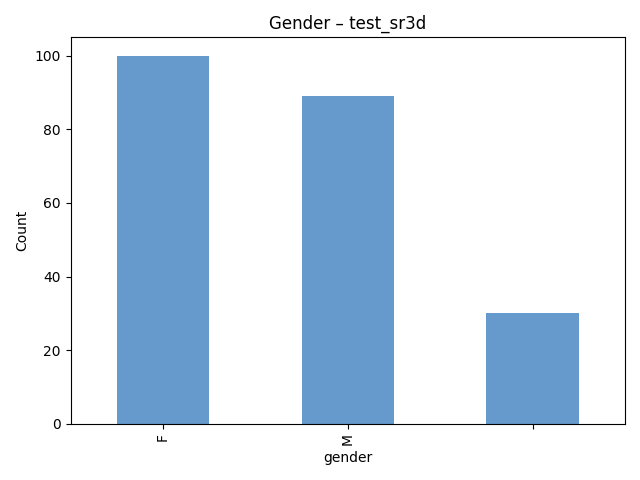}
    \caption{Gender distribution of the real-world Sr3D set.}
    \label{fig:gender_test_sr3d}
\end{figure}

\onecolumn
\textbf{Gender distribution}: Table \ref{fig:gender_all}, \ref{fig:gender_test_scanrefer}, \ref{fig:gender_test_nr3d} and \ref{fig:gender_test_sr3d} show the plots of gender distribution in the entire real-world dataset, ScanRefer set, Nr3D set and Sr3D set respectively.

\begin{figure}[h]
    \centering
    \includegraphics[width=0.8\linewidth]{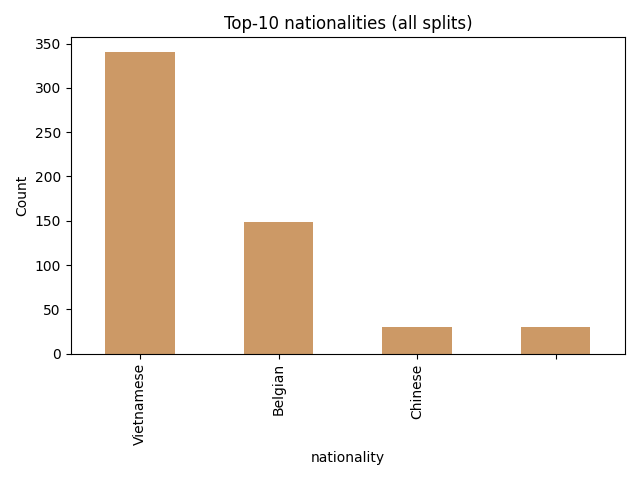}
    \caption{Nationality distribution of the entire real-world dataset, including ScanRefer, Nr3D and Sr3D.}
    \label{fig:nationality_all}
\end{figure}

\onecolumn
\textbf{Nationality distribution}: Table \ref{fig:nationality_all} shows the plot of nationality distribution in the entire real-world dataset.

\begin{figure}[h]
    \centering
    \includegraphics[width=0.8\linewidth]{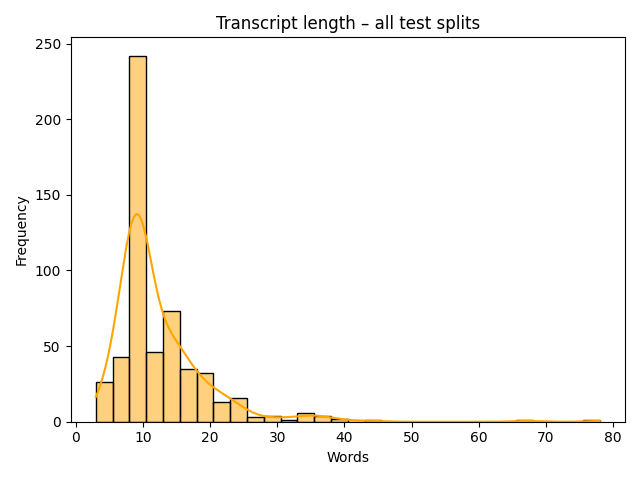}
    \caption{Wordcount distribution of the entire real-world dataset, including ScanRefer, Nr3D and Sr3D.}
    \label{fig:wordcount_all}
\end{figure}

\begin{figure}[h]
    \centering
    \includegraphics[width=0.8\linewidth]{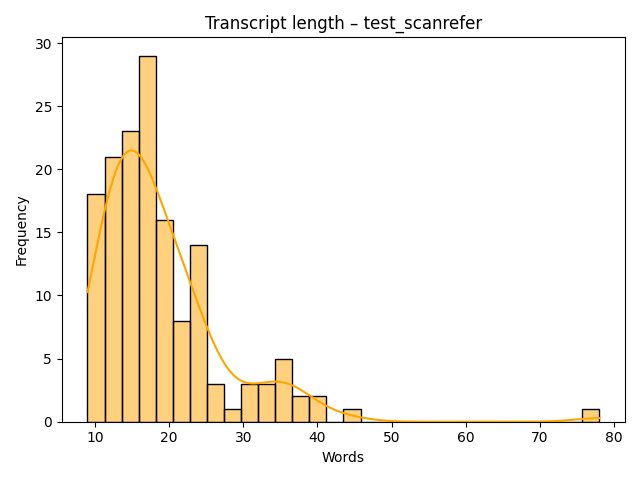}
    \caption{Wordcount distribution of the real-world ScanRefer set.}
    \label{fig:wordcount_test_scanrefer}
\end{figure}

\begin{figure}[h]
    \centering
    \includegraphics[width=0.8\linewidth]{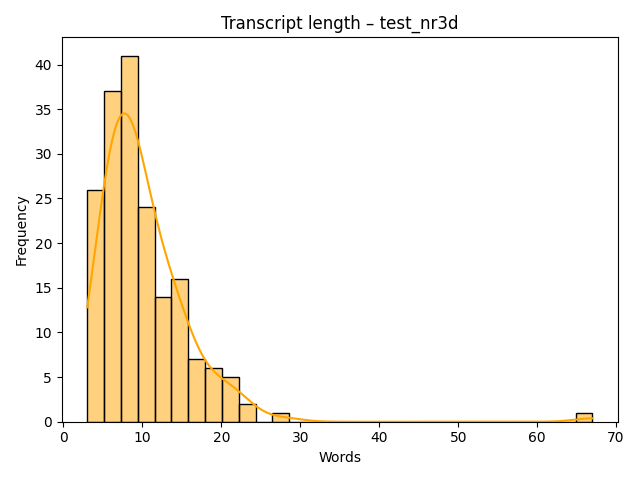}
    \caption{Wordcount distribution of the real-world Nr3D set.}
    \label{fig:wordcount_test_nr3d}
\end{figure}

\begin{figure}[h]
    \centering
    \includegraphics[width=0.8\linewidth]{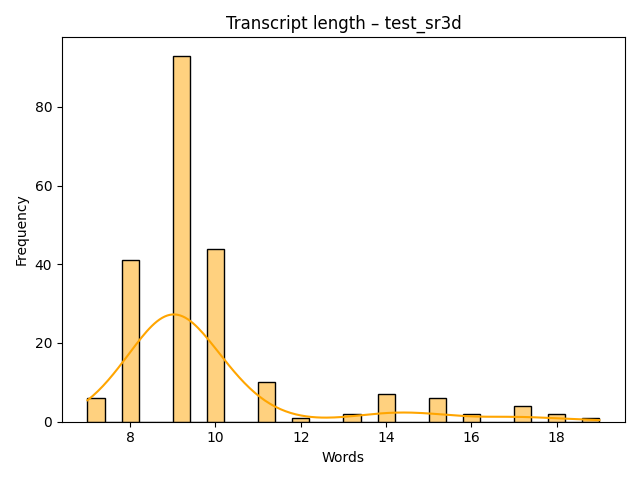}
    \caption{Wordcount distribution of the real-world Sr3D set.}
    \label{fig:wordcount_test_sr3d}
\end{figure}

\onecolumn
\textbf{Wordcount distribution}: Table \ref{fig:wordcount_all}, \ref{fig:wordcount_test_scanrefer}, \ref{fig:wordcount_test_nr3d} and \ref{fig:wordcount_test_sr3d} show the plots of wordcount distribution in the entire real-world dataset, ScanRefer set, Nr3D set and Sr3D set respectively.

\onecolumn
\section{Self-Supervised Speech Representation Learning}
\label{sec:SSL_speech}
This section aims to provide a comprehensive overview of self-supervised learning (SSL) behaviors for speech representation, which is the rationale to motivate our usage of wav2vec 2.0 in our multimodal fusion architecture.

\subsection{wav2vec 2.0 Architecture}
\begin{figure}[h]
    \centering
    \includegraphics[width=1.0\linewidth]{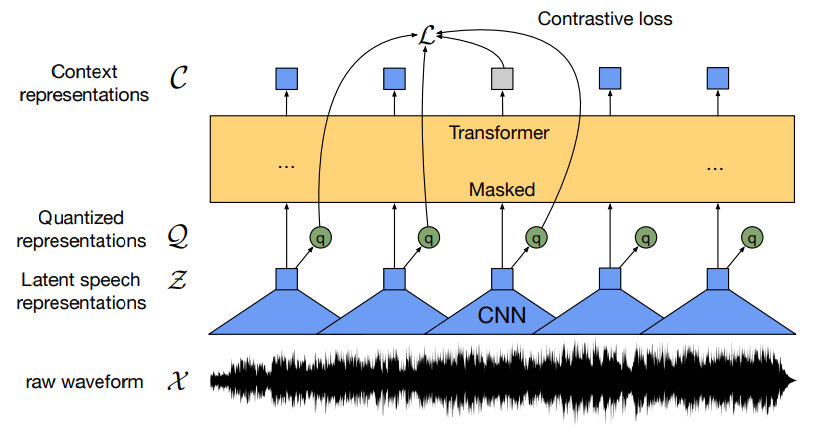}
    \caption{wav2vec 2.0 architecture}
    \label{fig:wav2vec2_architecture}
\end{figure}

wav2vec 2.0, introduced by \citet{Baevski2020wav2vec2A} (see Figure \ref{fig:wav2vec2_architecture}), is a SSL framework designed to learn speech representations directly from raw waveforms. The model enables data-efficient training of ASR systems by separating the pre-training and fine-tuning phases. The architecture consists of three key components: a feature encoder, a context network, and a quantization module.

\subsubsection{Self-Supervised Pre-training}
\textbf{Wave normalization}: The raw audio waveform $\mathcal{A}_T \in \mathbb{R}^T$ is first normalized to the range between 0 and 1 by the wave normalization function $\text{WaveNorm}$ before being pushed into the feature extractor, as shown in Equation \ref{Eq:WaveNorm}.
\begin{equation}
\label{Eq:WaveNorm}
\begin{split}
\mathcal{A}_T := \mathcal{A}_T^{\text{WaveNorm}} &= \text{WaveNorm}(\mathcal{A}_T)\\
&=\text{LayerNorm}(\mathcal{A}_T) \in \mathbb{R}^T
\end{split}
\end{equation}
$\text{WaveNorm}$ could be either layer normalization $\text{LayerNorm}$ \cite{ba2016layer} or batch normalization $\text{BatchNorm}$ \cite{ioffe2015batch}. 

\textbf{Feature Encoder}: The feature encoder is a multi-layer convolutional neural network (CNN) that processes raw audio input $\mathcal{A}_T \in \mathbb{R}^T$ to produce a sequence of latent audio representations $z_{a} \in \mathbb{R}^{{T'}_{a} \times d_{a}}$, where $T'_{a} < T$ due to down-sampling and $d_{a}$ is the feature dimension. Typically, the encoder consists of 7 convolutional layers with GELU activations \cite{hendrycks2016gaussian} and layer normalization \cite{ba2016layer}.

\begin{equation}
z_{a} = \text{FeatureEncoder}(\mathcal{A}_T)
\end{equation}

To be specific:
\begin{equation}
\begin{split}
z_{a} &= \text{FeatureEncoder}(\mathcal{A}_T)\\
       &= \text{FFW} \circ \text{CNNs} \circ \text{WaveNorm}(\mathcal{A}_T)
       \end{split}
\end{equation}

\textbf{Context Network}: The context network comprises a stack of Transformer encoder layers that model temporal dependencies in the latent feature sequence. These layers generate context-aware representations $c_{a} \in \mathbb{R}^{T'_{a} \times d_{a}}$ by applying self-attention and feedforward (FFW) operations.

\begin{equation}
c_{a} := \text{Transformer}(z_{a})
\end{equation}

Each Transformer block includes multi-head self-attention (MHSA), FFW sublayers, residual connections, and layer normalization. These allow the model to capture long-range dependencies in speech signals.

In an arbitrary $l$-th transformer layer, the output ${c_{a}}_{l}$ is briefly defined as:

\begin{equation}
\begin{split}
{c_{a}}_{l} &= \text{Transformer}({c_{a}}_{l-1})\\
&=\text{FFW} \circ \text{MHSA}({c_{a}}_{l-1})
\end{split}
\end{equation}
where $\text{MHSA}$ is multi-head attention which is a function defined by self-attention functions $\text{SA}$:

\begin{equation}
\text{MHSA}({c_{a}}_{l-1}) = \text{SA}({c_{a}}_{l-1}) + {c_{a}}_{l-1}
\end{equation}

Then, we have a full equation for an arbitary $l$-th Transformer layer:
\begin{equation}
\begin{split}
{c_{a}}_{l} &= \text{FFW}(\text{MHSA}({c_{a}}_{l-1})) + \text{MHSA}({c_{a}}_{l-1}) \\
&= \text{FFW}(\text{SA}({c_{a}}_{l-1}) + {c_{a}}_{l-1}) + \left [\text{SA}({c_{a}}_{l-1}) + {c_{a}}_{l-1}\right ]
\end{split}
\end{equation}

For layer-wise formulation, the 0-th Transformer layer (the first layer) is connected to the feature encoder, which is defined as:
\begin{equation}
{c_{a}}_{0} = \text{Transformer}(z_{a})
\end{equation}

Given an $L$-Transformer-layer wav2vec 2.0 architecture, the $L-1$-th Transformer layer (the final layer) is defined as a chain function as:
\begin{equation}
\begin{split}
{c_{a}}_{L-1} &= \text{Transformer}({c_{a}}_{L-2}) \\
&= \text{Transformer} \circ \text{Transformer} \circ ... \circ \text{Transformer}(x^{\tau}_{0})\\
&= \text{Transformer} \circ \text{Transformer} \circ ... \circ \text{Transformer} \circ \text{Transformer}(z_{a})
\end{split}
\end{equation}
where $L$ is the total number of Transformer layers in the encoder, layer indices start from $0$ to $L-1$.

\textbf{Quantization Module}: To formulate a contrastive learning task, the model discretizes the latent features $z_{a}$ into quantized targets $q_{t_{a}} \in \mathbb{R}^d_{a}$ using a quantization module. This module employs Gumbel-softmax-based vector quantization with multiple codebooks.

Let $G_{a}$ be the number of codebooks and $V_{a}$ the number of entries per codebook. Each quantized representation is obtained as:

\begin{equation}
q_{t_{a}} = \text{Concat}(e_{g_{1_{a}}}, e_{g_{2_{a}}}, \ldots, e_{g_{G_{a}}})
\end{equation}

where each $e_{g_{i_{a}}} \in \mathbb{R}^{d_{a}/G_{a}}$ is a learned embedding vector selected from the $i$-th codebook.

\textbf{Pre-training Objective}: During pre-training, the model masks a subset of the latent features and uses the context representations to identify the corresponding quantized targets from a pool of negatives. The primary learning signal is a contrastive loss defined as:

\begin{equation}
\mathcal{L}_{\text{contrastive}} = -\log \frac{\exp(\text{sim}(c_{t_{a}}, q_{t_{a}}) / \kappa_{a})}{\sum\limits_{q'_{a} \in \mathcal{Q}_{t_{a}}} \exp(\text{sim}(c_{t_{a}}, q'_{a}) / \kappa_{a})}
\end{equation}

where $\text{sim}(\cdot, \cdot)$ is cosine similarity, $\kappa_{a}$ is a temperature hyperparameter, and $\mathcal{Q}_{t_{a}}$ contains one true quantized vector and multiple negatives.

To encourage diversity across codebook entries, a diversity loss is added:

\begin{equation}
\mathcal{L}_{\text{diversity}} = \frac{G_{a}}{V_{a}} \sum_{g_{a}=1}^{G_{a}} \sum_{v_{a}=1}^{V_{a}} p_{g_{a},v_{a}} \log p_{g_{a},v_{a}}
\end{equation}

where $p_{g,v}$ is the average selection probability of the $v$-th entry in the $g$-th codebook. The total pre-training loss is:

\begin{equation}
\mathcal{L}_{a} = \mathcal{L}_{\text{contrastive}} + \alpha_{a} \cdot \mathcal{L}_{\text{diversity}}
\end{equation}

with $\alpha_{a}$ being a tunable weight.

\subsubsection{Supervised Fine-tuning}
Once pre-trained, the context representations $c_{a}$ are used for supervised ASR by appending a randomly initialized linear projection layer and training the model with a Connectionist Temporal Classification (CTC) loss \cite{graves2006connectionist, graves2012connectionist}. The entire model is fine-tuned end-to-end using a small amount of labeled data, enabling high accuracy even with limited supervision.

See Section \ref{sec:ASR} for details of supervised fine-tuning ASR.

\subsection{Rationale}
The rapid advancement of deep learning has revolutionized the field of speech processing, enabling significant improvements in tasks such as ASR, speaker identification, emotion recognition, and speech synthesis. Traditionally, these tasks have relied heavily on supervised learning \cite{synnaeve2020end, luscher2019rwth, zeyer2019:trafo-vs-lstm-asr, zeyer2020:transducer}, which requires large volumes of labeled data. However, obtaining high-quality labeled speech data is both expensive and time-consuming, especially when considering the wide variability in languages, accents, recording conditions, and speaker characteristics. This has led to a growing interest in SSL, an approach that leverages vast amounts of unlabeled data to learn meaningful representations without explicit annotations \cite{baevski2020effectiveness, fan2022towards, mohamed2022self, vieting2023efficient}.

Self-supervised speech representation learning aims to extract high-level, informative features from raw audio signals by solving pretext tasks derived from the inherent structure of the data. These pretext tasks are designed such that solving them requires understanding relevant patterns in the speech signal, such as phonetic content, prosody, or speaker identity \cite{schneider2019wav2vec, xu2021explore, baevskivq}. Once trained, the resulting representations can be fine-tuned or directly applied to downstream tasks with minimal supervision, significantly reducing the dependence on labeled data \cite{fan2024benchmarking, liu2021tera, lei2024mix}.

Recent breakthroughs in SSL, particularly inspired by advances in natural language processing (e.g., BERT \cite{devlin2019bert}, GPT \cite{openai2024gpt4technicalreport, brown2020languagemodelsfewshotlearners, radford2019language}) and computer vision (e.g., SimCLR \cite{chen2020simple}, MoCo \cite{he2020momentum}), have led to the development of powerful speech models such as wav2vec \cite{schneider2019wav2vec}, HuBERT \cite{hsu2021hubert}, and WavLM \cite{chen2022wavlm}. These models have demonstrated state-of-the-art performance across a wide range of speech-related benchmarks, often outperforming fully supervised counterparts when only limited labeled data is available \cite{Baevski2020wav2vec2A, schneider2019wav2vec}. Moreover, SSL has opened new avenues for learning more generalizable, robust, and multilingual representations \cite{bai2022joint, fu2022losses, sakti2023leveraging}.

\begin{figure}[h]
    \centering
    \includegraphics[width=0.8\linewidth]{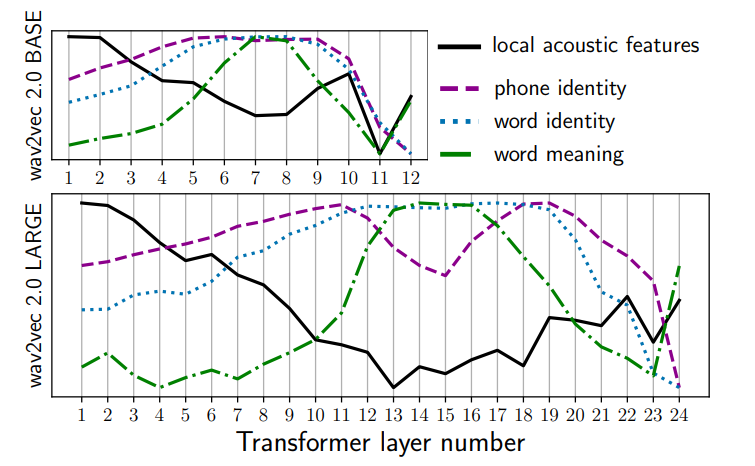}
    \caption{Visualization of properties encoded at different wav2vec 2.0 layers.}
    \label{fig:layerwise_w2v2_visualization}
\end{figure}

The wav2vec 2.0 transformer exhibits an autoencoder-like behavior \cite{vielzeuf2024investigating, chiu2022self}: the representations start deviating from the input speech features followed by a reverse trend where even deeper layers become more similar to the input, as if reconstructing the input. 
\begin{enumerate}
    \item The layer-wise progression of representations exhibits an acoustic-to-linguistic hierarchy: lower layers encode acoustic features, followed sequentially by phonetic, word identity, and semantic information, before reversing this trend in the upper layers, as shown in Figure \ref{fig:layerwise_w2v2_visualization}.
    \item ASR fine-tuning disrupts this autoencoder-like behavior in the upper layers, enhancing their capacity to encode lexical information.
    \item The initial transformer and final CNN layers show high correlation with mel spectrograms, indicating convergence toward human-engineered features.
    \item The SSL model encodes some semantic content.
    \item The final two layers often deviate from preceding patterns
\end{enumerate}

\subsection{Analysis Methods}
\subsubsection{Canonical Correlation Analysis}

Canonical Correlation Analysis (CCA) \cite{hotelling1936relations} is a classical statistical technique designed to quantify the linear relationships between two multivariate random variables. Given two sets of continuous-valued random vectors, CCA identifies pairs of canonical directions—one for each set—such that the correlation between the projections of the vectors onto these directions is maximized. This results in a sequence of canonical correlation coefficients that capture the degree of linear alignment between the two representational spaces.

In the context of SSL models such as wav2vec 2.0, CCA has proven to be a valuable tool for analyzing the internal structure of learned representations. wav2vec 2.0 encodes raw audio waveforms into hierarchical feature representations through a series of convolutional and Transformer layers. By applying CCA, we can quantify the representational similarity across layers of the model, offering insight into how acoustic and linguistic information is progressively abstracted.

\citet{pasad2021layer} employ CCA in two complementary ways. First, they compute pairwise CCA scores between different layers of the wav2vec 2.0 Transformer encoder to investigate the evolution and redundancy of learned features. This helps assess whether certain layers exhibit similar information encoding patterns, or whether deeper layers introduce significant representational shifts.

Second, \citet{pasad2021layer} apply CCA to measure the similarity between the internal layer representations of wav2vec 2.0 and external reference vectors. These reference vectors include pre-trained word embeddings (e.g., Word2Vec \cite{mikolov2013distributed} or GloVe \cite{pennington2014glove}) and low-level acoustic features (e.g., Mel-frequency cepstral coefficients or log-Mel spectrograms). This cross-modal comparison enables us to determine the extent to which specific Transformer layers align with either phonetic-level acoustic information or semantically-rich linguistic abstractions. Through this analysis, we gain deeper interpretability into how wav2vec 2.0 encodes and transitions between speech and language representations \cite{morcos2018insights, kornblith2019similarity, raghu2017svcca}.

\subsubsection{Mutual Information Estimation}
While CCA is a natural choice for quantifying relationships between pairs of continuous-valued vector representations, it is limited to capturing linear correlations and does not generalize well to the dependence between learned representations and categorical linguistic units such as phones or words. Instead, \citet{pasad2021layer} adopt \textit{mutual information (MI)} as a more general measure of statistical dependence between the latent representations $\mathbf{y}_{\text{phn}}$ or $\mathbf{y}_{\text{wrd}}$—which are extracted from intermediate layers of the wav2vec 2.0 model—and their corresponding ground-truth phoneme or word labels. 

Since the model outputs continuous-valued representations, \citet{pasad2021layer} follow prior work \cite{Baevski2020wav2vec2A, alain2016understanding} and discretize them using clustering (e.g., $k$-means), thereby enabling estimation of mutual information via co-occurrence statistics.

The resulting MI metrics, denoted as \textbf{MI-phone} and \textbf{MI-word}, quantify the amount of phonetic or lexical information preserved in the internal feature representations. Higher MI indicates a stronger correlation between learned representations and linguistic targets, providing insight into the degree of linguistic abstraction encoded by the model during SSL.

\subsection{Findings of Self-Supervised Representation Learning}
\subsubsection{Reconstruction Behavior}
\begin{figure}[h]
    \centering
    \includegraphics[width=0.7\linewidth]{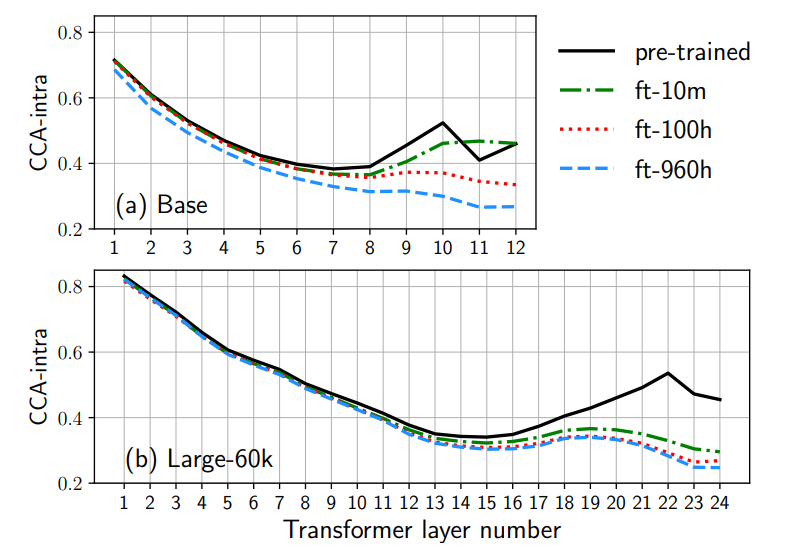}
    \caption{CCA similarity with local features}
    \label{fig:CCA_similarities}
\end{figure}
Figure \ref{fig:CCA_similarities} presents a comparison of transformer layer representations with the local features extracted by the CNN module (layer 0), using CCA similarity. The pre-trained model (solid black curve) exhibits an autoencoder-like pattern: representations initially diverge from the input features with increasing depth, but subsequently reconverge in deeper layers, indicating a reconstruction-like behavior. This trend is disrupted in the final two layers (see Section below). Given that the training objective involves distinguishing a masked input segment from distractors, it is expected that the final layers encode representations similar to the input. A comparable pattern—termed context encoding and reconstruction—has been previously observed in BERT for masked language modeling objectives \cite{voita2019bottom}.
  
\subsubsection{Encoded Acoustic-Linguistic Information}
\citet{pasad2021layer} analyzed how specific properties are encoded across different model layers. It is important to note that all experiments are conducted using features extracted from short temporal spans, corresponding to frame-, phone-, or word-level segments. Any observed increase in the amount of encoded "information" across layers for these local representations can be attributed to the contextualization enabled by the self-attention mechanism, which allows each frame-level output to incorporate information from the entire utterance. Conversely, a reduction in localized "information" across layers may result from de-localization, wherein the representation becomes increasingly distributed and less confined to the original temporal segment.

\begin{figure}[h]
    \centering
    \includegraphics[width=0.7\linewidth]{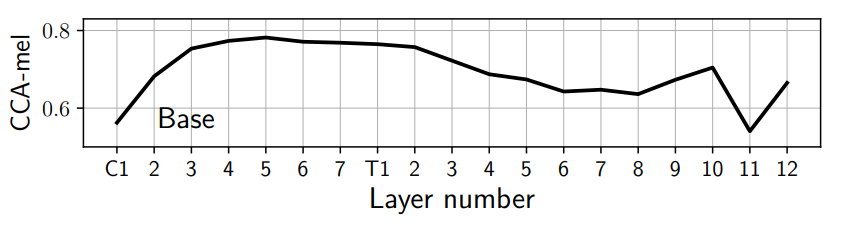}
    \caption{CCA similarity between layer representations and
fbank; Ci: CNN layer i, Tj: transformer layer j.}
    \label{fig:frame_level_acoustic_content}
\end{figure}

\textbf{Frame-level acoustic information}: Figure \ref{fig:frame_level_acoustic_content} presents the layer-wise CCA similarity between filterbank (fbank) features and the representations from the wav2vec 2.0 Base model. In the initial layers, the correlation increases progressively with depth. A similar trend is observed for the Large models, which exhibit high CCA values (> 0.75) between layers C4 and T2. These results suggest that the model implicitly learns representations analogous to fbank features, indicating the potential for simplifying wav2Vec 2.0 by directly using fbank inputs. However, to our best knowledge, the potential suggested by \citet{pasad2021layer} has not been empirically proven yet.

\begin{figure}[h]
    \centering
    \includegraphics[width=0.7\linewidth]{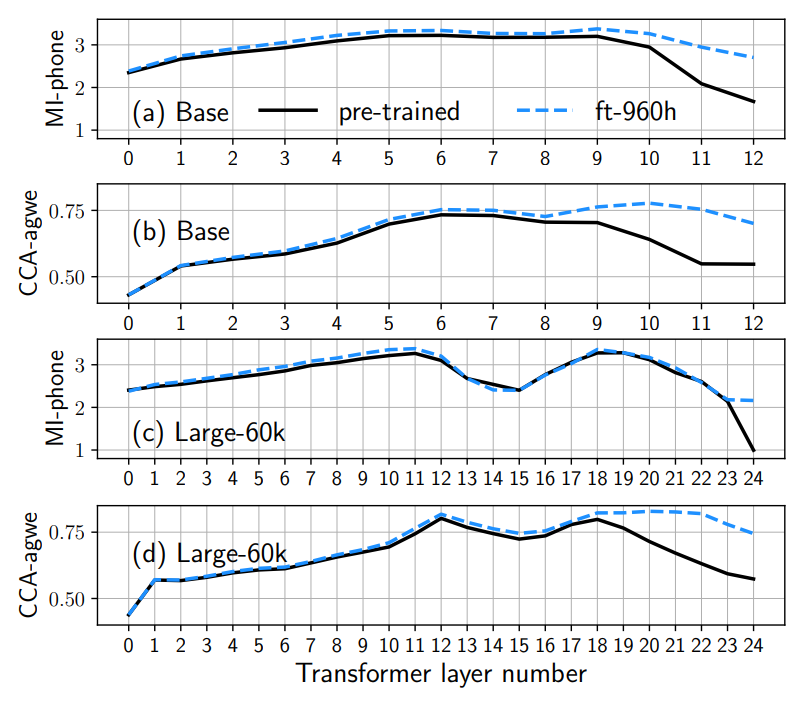}
    \caption{MI with phone labels (max: 3.6) and CCA similarity
with AGWE.}
    \label{fig:phonetic_information}
\end{figure}

\textbf{Phonetic information}: \citet{pasad2021layer} quantify the phonetic information encoded in the pre--trained model using two metrics: mutual information with phone labels (MI-phone) and canonical correlation analysis with AGWEs (CCA-agwe), as visualized in Figure \ref{fig:phonetic_information}. Given that AGWEs are designed to represent phonetic content, the similarity in trends between the MI-phone and AGWE curves supports this expectation. In the wav2vec 2.0 Base model, phonetic information peaks around layers 6–7. We, to the best of our knowledge, found this behavior consistent with prior findings \cite{hsu2021hubert_analysis} which analyzed the behavior of HuBERT \cite{hsu2021hubert}. In contrast, the Large-60k model exhibits prominent phonetic encoding at layers 11 and 18/19, with a notable decline in intermediate layers.

\begin{figure}[h]
    \centering
    \includegraphics[width=0.7\linewidth]{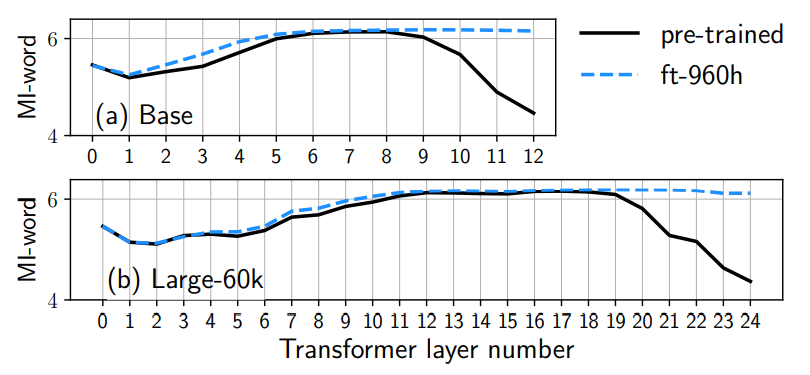}
    \caption{MI with word labels (max: 6.2).}
    \label{fig:word_identity}
\end{figure}

\textbf{Word identity}: Figure \ref{fig:word_identity} presents the MI between layer representations and word labels. For the wav2vec 2.0 Base model, the observed trends resemble those of MI with phone labels (Figure \ref{fig:phonetic_information}). In the Large-60k model (Figure \ref{fig:word_identity}), word identity is consistently encoded across layers 12 to 18, without the decline observed in the MI-phone curve. This behavior shows that, to the best understanding of \citet{pasad2021layer}'s work, MI-word and word discrimination are always highly correlated.

\subsubsection{Word Meaning Representation}
\begin{figure}[h]
    \centering
    \includegraphics[width=0.7\linewidth]{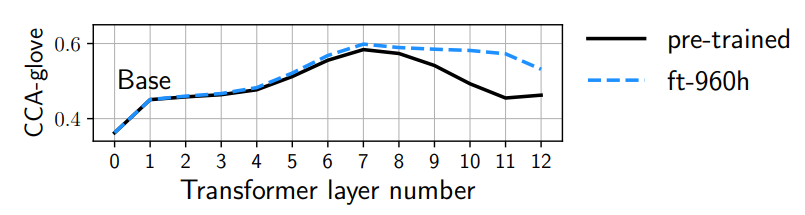}
    \caption{CCA similarity with GloVe embeddings \cite{pennington2014glove}.}
    \label{fig:word_meaning_representation}
\end{figure}

Although certain linguistic features appear critical for the model to solve the SSL objective, it remains unclear whether semantic content—specifically word meaning—is among them. To investigate this, \citet{pasad2021layer} assess the encoding of word meaning in wav2Vec 2.0 by computing the CCA similarity between word segment representations and GloVe embeddings \cite{pennington2014glove}, as illustrated in Figure \ref{fig:word_meaning_representation}. The results indicate that the middle layers—layers 7–8 in the Base model and 14–16 in the Large-60k model—encode the richest contextual information. Notably, the narrower plateau of peak performance in these curves compared to the MI curves in Figure \ref{fig:word_identity} suggests that central layers are more specialized in capturing semantic content, whereas peripheral layers primarily encode lower-level linguistic features without semantic abstraction.

\subsubsection{Fine-tuning Effect}
\begin{figure}[h]
    \centering
    \includegraphics[width=0.7\linewidth]{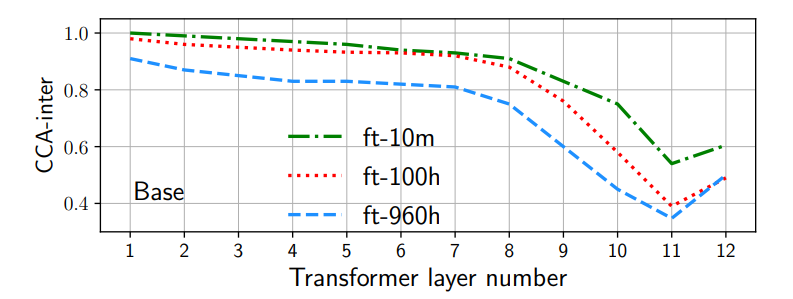}
    \caption{CCA similarity between each layer of a pre-trained
model and the same layer of fine-tuned models.}
    \label{fig:finetuning_effect}
\end{figure}

As shown in Figure \ref{fig:CCA_similarities} (CCA-intra), fine-tuning disrupts the autoencoder-like behavior of the model. Post fine-tuning for ASR, the deeper layers, which previously aimed to reconstruct the input, increasingly diverge from it, indicating a shift toward learning task-specific representations. Additionally, Figure \ref{fig:finetuning_effect} reveals that the upper layers undergo the most significant changes during fine-tuning, implying that the pre-trained model may provide suboptimal initialization for these layers in ASR tasks. This observation, to the best of our knowledge, aligns with findings in BERT language modelling \cite{zhangrevisiting}, where re-initialization of top layers prior to fine-tuning improves performance.

The results also suggest that fine-tuning with character-level CTC loss \cite{graves2006connectionist} is more strongly associated with encoding word identity than phone identity, as anticipated.

We observed that the final layers of wav2Vec 2.0 undergo the most substantial modifications during fine-tuning (Figure \ref{fig:finetuning_effect}) and exhibit reduced encoding of linguistic information relevant to ASR. These findings suggest that certain upper layers may offer suboptimal initialization for downstream ASR tasks.

\onecolumn
\section{Weakly Supervised Speech Representation Learning}
\subsection{Attention Encoder Decoder (AED)}

\begin{figure*}[h]
    \centering
    \includegraphics[width=1.0\linewidth]{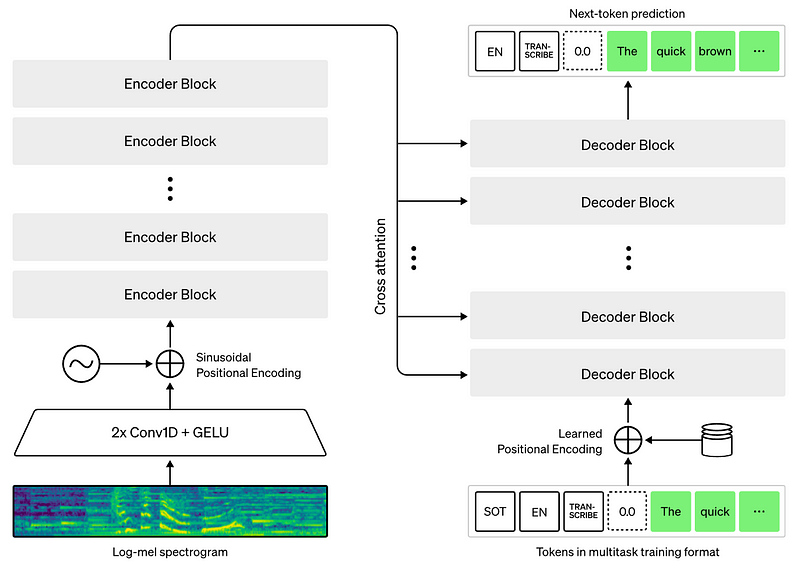}
    \caption{OpenAI's Whisper architecture. Whisper is a Transformer-based AED architecture, using MFCC features as input instead of raw waveform.}
    \label{fig:whisper_architecture}
\end{figure*}

\begin{figure*}[h]
    \centering
    \includegraphics[width=1.0\linewidth]{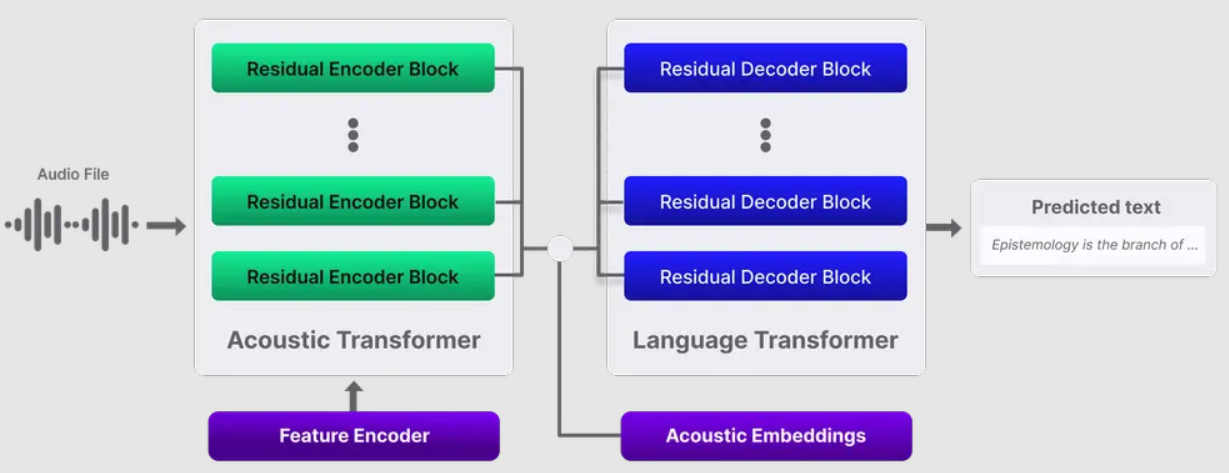}
    \caption{Deepgram's Nova-2 architecture. To our best understanding of Deepgram's documentation, Deepgram's Nova-2 is a Transformer-based AED architecture, using raw waveform as input instead of MFCC like Whisper. Feature extraction from raw waveform is probably  conducted by a learnable feature encoder, e.g. a block of CNNs like wav2vec 2.0. Between encoder-decoder space, (unknown) acoustic embeddings are probably added as cross-attention.}
    \label{fig:deepgram_architecture}
\end{figure*}

As for AED models, Whisper architecture is shown in Figure \ref{fig:whisper_architecture}, and Deepgram architecture is shown in Figure \ref{fig:deepgram_architecture}.

An ASR model is used to transcribe speech into text by mapping an acoustic vector $x^{T}_{1} := x_{1}, x_{2}, ..., x_{T}$ of length $T$ to the most likely word sequence $w^{N}_{1}$ of length $N$. The word sequence probability is described as:

\begin{equation}
\label{eq:word_sequence_general}
p(w_{1}^{N}|x_{1}^{T}) = \prod_{n=1}^{N} p(w_n|w_{1}^{n-1},x_{1}^{T}).    
\end{equation}

In the ASR encoder-decoder architecture, given $D$ as the feature dimension size, the input audio signal matrix could be described as $x^{T}_{1} \in \mathbb{R}^{T \times D_{input}}$. When simplified, downsampling before or inside the encoder - conducted by a fixed factor, such as striding in a CNN - is removed. Thus, the encoder output sequence is as follows:

\begin{equation}
h_{1}^{T} = Encoder(x_{1}^{T}) \in \mathbb{R}^{T \times D_{encoder}}. 
\end{equation}

Using a stack of \text{Transformer} blocks \cite{vaswani2017attention}, the encoder output sequence is described as function composition:
\begin{equation}
h_{1}^{T} = {\text{Transformer}}_{0} \circ ... \circ {\text{Transformer}}_{N_{EncLayers}}(x_{1}^{T}).
\end{equation}

In the decoder, the probability for each single word is defined as:

\begin{equation}
\begin{split}
p(w_n|w_{1}^{n-1},x_{1}^{T}) &= p(w_n|w_{1}^{n-1},h_{1}^{T}(x_{1}^{T}))\\
&= p(w_n|w_{1}^{n-1},h_{1}^{T}). 
\end{split}   
\end{equation}

Based on Equation \ref{eq:word_sequence_general}, the word sequence probability given the output of encoder is described as:
\begin{equation}
p(w_{1}^{N}|x_{1}^{T}) = \prod_{n=1}^{N} p(w_n|w_{1}^{n-1},h_{1}^{T}).    
\end{equation}

Then, decoder hidden state is formulated as:
\begin{equation}
g_n = \mathcal{F}(g_{n-1},w_{n-1}, c_n) \in \mathbb{R}^{D_{g}},
\end{equation}
where $\mathcal{F}$ is neural network; $D_{g}$ is hidden state dimension; and $c_n$ is context vector, e.g. weighted sum of encoder outputs via attention mechanism.

The attention mechanism in the decoder is described via 3 components: context vector $c_n$, attention weights $\alpha_{n,t}$, and attention energy $e_{n,t}$:

\begin{equation}
\begin{split}
c_n &= \sum_{t=1}^{T} \alpha_{n,t} {h}_{t} \in \mathbb{R}^{D_{encoder}},\\
\alpha_{n,t} &= \frac{\exp(e_{n,t})}{\sum_{t'=1}^{T}\exp(e_{n,t'})} \\
&= Softmax_{T}(\exp(e_{n,t})) \in \mathbb{R},\\
e_{n,t} &= Align(g_{n-1}, h_t) \in \mathbb{R} \\
&= W_{2} \cdot \tanh(W_{1} \cdot [g_{n-1}, h_t]),
\end{split}
\end{equation}
where $n$ is decoder step; $t$ is encoder frame; $\alpha \in \mathbb{R}^{T \times N}$ is attention weight matrix; $\alpha_n \in \mathbb{R}^{T}$ is normalized probability distribution over $t$; $Softmax_{T}$ is Softmax function over spatial dimension $T$, not feature dimension; $W_{1} \in \mathbb{R}^{(D_{g}+D_{encoder}) \times D_{key}}$; $W_{2} \in \mathbb{R}^{D_{key}}$.

In the decoding, the output probability distribution over vocabulary is defined as:
\begin{equation}
\begin{split}
&p(w_{n} = *|w_{1}^{n-1}, h_{1}^{T})\\
&= Softmax(MLP(w_{n-1}, g_n, c_n)) \in \mathbb{R}^{N}, 
\end{split}
\end{equation}
where $MLP$ is Multi-layer Perceptron.

To train an AED model,  sequence-level frame-wise cross-entropy loss is employed:
\begin{equation}
\begin{split}
\mathscr{L}_{AED} &= - \sum_{(x_{1}^{T}, w_{1}^{N})} \log p(w_{1}^{N}|x_{1}^{T})\\
&= - \sum_{(x_{1}^{T}, w_{1}^{N})} \sum_{n=1}^{N} \log p(w_n|w_{1}^{n-1},x_{1}^{T}).
\end{split}
\end{equation}

During beam search, the auxilary quantity for each unknown partial string (tree of partial hypotheses) $w_{1}^{n}$ is defined as:
\begin{equation}
\begin{split}
Q(n; w_{1}^{n}) :&= \prod_{n'=1}^{n} p(w_{n'}|w_{0}^{n'-1},x_{1}^{T})\\
&= p(w_{n}|w_{0}^{n-1},x_{1}^{T}) \cdot Q(n-1, w_{1}^{n-1}).
\end{split}
\end{equation}

After discarding the less likely hypotheses in the beam search, the word sequence probability is calculated by the best hypothesis:
\begin{equation}
p(w_{1}^{N}|x_{1}^{T}) = Q(N; w_{1}^{N}).    
\end{equation}

\onecolumn
\section{Raw Waveform and Feature Extraction}
\subsection{Mel-Frequency Cepstral Coefficients (MFCC)}
\begin{figure*}[h]
    \centering
    \includegraphics[width=1\linewidth]{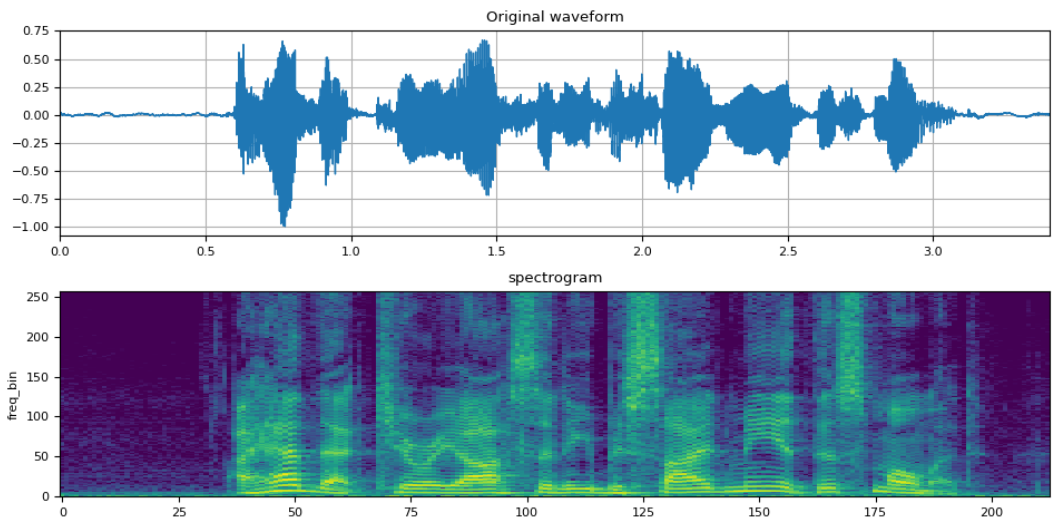}
    \caption{MFCC visualization. The computation of MFCCs begins by dividing the original waveform into overlapping 20ms frames.}
    \label{fig:MFCC_visualization}
\end{figure*}

MFCC serves as a compact representation of the audio signal's spectral properties. The computation of MFCCs begins by dividing the input signal $x^{T}_{1} := x_{1}, x_{2}, ..., x_{T}$ into overlapping frames, as visualized in Figure \ref{fig:MFCC_visualization}\footnote{\citet{golik2020data}'s Dissertation at RWTH Aachen University described MFCC more comprehensively.\\
MFCC visualization image is retrieved from Pytorch library.}.

\textbf{Pre-emphasis}: The audio signal, sampled at 16 kHz with a step size of 10 ms, is processed by extracting 160 consecutive samples from the Pulse Code Modulation (PCM) waveform for each frame. These 10 ms frames are non-overlapping, ensuring that stacking adjacent vectors avoids discontinuities. The 16-bit quantized samples, which span the integer range from $-2^{15}$ to $+2^{15}$, must be normalized to a numerically stable range. This normalization is achieved by applying mean and variance normalization, either globally across the entire training dataset or on a per-utterance basis. A commonly employed processing technique, known as high-frequency pre-emphasis, can be implemented by computing the differences between adjacent samples, as illustrated below:

\begin{equation}
\mathbcal{x}'_t = \mathbcal{x}_t - \mathbcal{x}_{t-1} \in \mathbb{R}
\end{equation}

A sequence of $16 \, \text{kHz} \times 10 \, \text{ms} = 160$ pre-emphasized waveform samples can then be considered a feature vector:

\begin{equation}
\hat{\mathbcal{x}}_t = {\mathbcal{x}'}^{t}_{t-160+1} \in \mathbb{R}^{160}
\end{equation}

\textbf{Amplitude spectrum - FFT}: The short-time Fourier transform (STFT) is applied to overlapping windows with a duration of \( 25 \, \text{ms} \). Given a sampling rate of \( 16 \, \text{kHz} \), this window length corresponds to $25 \, \text{ms} \times 16 \, \text{kHz} = 400 \, \text{samples}$. To facilitate computation using the fast Fourier transform (FFT), the sample count is zero-padded to the next power of two, resulting in \( 2^9 = 512 \).
\begin{equation}
\begin{split}
&\mathbcal{z}_t \in \mathbb{R}^{512} \\
&=\begin{bmatrix}
\mathbcal{x}^{t'}_{t-400+1} & \mathbcal{x}^{t'}_{t-400+2} & \dots & \mathbcal{x}^{t'}
\underbrace{0 \dots 0}_{\text{zero-padding}}
\end{bmatrix}
\end{split}
\end{equation}

The extended sample vector is weighted using a Hann window, which exhibits smaller side lobes in the amplitude spectrum compared to a rectangular window:
\begin{equation}
\begin{split}
\mathbcal{w}^{(n)} &= 0.5 - 0.5 \cos\left(\frac{2\pi (n - 1)}{512 - 1}\right), \\
&\quad 1 \leq n \leq 512
\end{split}    
\end{equation}

\begin{equation}
    \mathbcal{s}_t^{(n)} = \mathbcal{z}_t^{(n)} \cdot \mathbcal{w}^{(n)}
\end{equation}

While the discrete STFT could be done directly by evaluating the sum 
\begin{equation}
\begin{split}
\mathbcal{S}_t^{(\mathbb{F})} &= \sum_{n=0}^{512-1} \mathbcal{s}_t^{(n)} \cdot \exp\left(-j \frac{2\pi}{512}\mathbb{F}n\right),\\
&\quad 1 \leq \mathbb{F} \leq 512
\end{split}
\end{equation}
the complexity can be reduced from \(\mathcal{O}(N^2)\) to \(\mathcal{O}(N \log N)\) by applying the fast Fourier transform. 

The 512-FFT results in a 257-dimensional vector because of the symmetry of the amplitude spectrum of a real-valued signal. The phase spectrum is removed. 
\begin{equation}
\begin{split}
\hat{\mathbcal{x}}_t &= 
\begin{bmatrix}
|\mathbcal{S}_t^{(0)}| & |\mathbcal{S}_t^{(1)}| & \dots & |\mathbcal{S}_t^{(512/2)}| 
\end{bmatrix} \\
&\in \mathbb{R}^{512/2+1}
\end{split}
\end{equation}

\textbf{MFCC}: The MFCC feature extraction is based on the STFT of the pre-emphasized speech signal \cite{davis1980comparison}. It considers the nonlinear sensitivity of human auditory perception to variations in frequency. This is evidenced that the filter bank used to integrate the magnitude spectrum $|\mathbcal{S}^{(\mathbb{F})}_t|$ consists of $\mathbb{I}$ filters equidistantly spaced on the mel scale. The mel scale is a logarithmically scaled frequency axis. The $k$-th frequency bin of the FFT centered around $\mathbb{F}_k$ Hz is then mapped to \( \tilde{\mathbb{F}}_k \) on the mel scale:
\begin{equation}
\mathbb{F}_k = \frac{k}{512} \cdot \mathbb{F}_\mathbcal{s}    
\end{equation}
\begin{equation}
\tilde{\mathbb{F}}_k = 2595 \cdot \log_{10}\left(1 + \frac{\mathbb{F}_k}{700 \text{ Hz}} \right)     
\end{equation} 

The filter center \( \tilde{\mathbb{F}}^{(i)}_c \) of the \( i \)-th triangular filter is then placed at \( i \cdot \tilde{\mathbb{F}}_b \), where the bandwidth \( \tilde{\mathbb{F}}_b \) corresponds to \( \tilde{\mathbb{F}}_{512} / \mathbb{I} \). With these parameters, the coefficients of the \( i \)-th triangular filter can be calculated explicitly as a piecewise linear function and stored in a weight vector \( \mathbcal{v}_i \in \mathbb{R}^{N/2+1} \). 

By applying discrete cosine transform (DCT), the MFCC features are extracted from the logarithm filter outputs:
\begin{equation}
\mathbcal{X}^{(i)}_t = \log_{10}\left( \sum_{\mathbb{F}=0}^{512} |\mathbcal{S}^{(\mathbb{F})}_t| \mathbcal{v}^{(\mathbb{F})}_i \right)    
\end{equation}
\begin{equation}
\mathbcal{c}_{m,i} = \cos \left( \frac{\pi m (i + 0.5)}{\mathbb{I}} \right)   
\end{equation}
\begin{equation}
\mathbcal{C}^{(m)}_t = \sum_{i=0}^{\mathbb{I}-1} \mathbcal{c}_{m,i} \mathbcal{X}^{(i)}_t    
\end{equation}
\begin{equation}
\hat{\mathbcal{x}}_t = \left[ \mathbcal{C}^{(0)}_t \mathbcal{C}^{(1)}_t \dots \mathbcal{C}^{(\mathbb{I}-1)}_t \right] \in \mathbb{R}^\mathbb{I}    
\end{equation}

\subsection{SpecAugment}
SpecAugment \cite{park2019specaugment} is a data augmentation technique for ASR that manipulates spectrograms to improve model robustness by randomly applying masking in consecutive frames in the time axis as well as consecutive dimensions in the feature axis. It performs three main transformations\footnote{\citet{bahar2019using} analyzed deeply in end-to-end speech translation. \citet{park2019specaugment} stated that time warping is the most expensive
and the least influential, we do not include it here.}: time warping, frequency masking, and time masking.

\begin{figure}[h]
    \centering
    \includegraphics[width=1.0\linewidth]{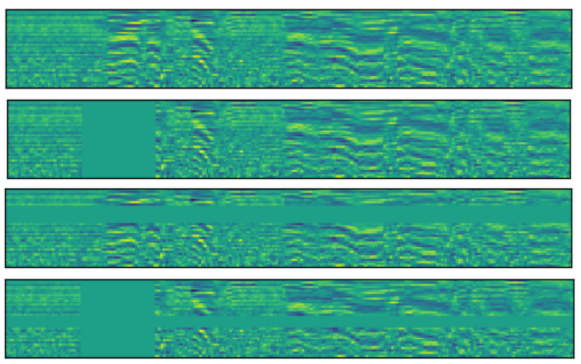}
    \caption{SpecAugment visualization. From top to bottom, the figures show the spectrogram of the input audio with no data augmentation, time masking, frequency masking and both masking applied.}
    \label{fig:specaug}
\end{figure}

Figure \ref{fig:specaug} shows examples of the individual augmentations applied to a single input. 

\textbf{Time Masking}: Given an audio signal $x^{T}_{1} := x_{1}, x_{2}, ..., x_{T}$ of length $T$. Time masking is masking of $\text{\textturntwo}$ successive time steps $[t, t + \tau)$, where we set:
\begin{equation}
(x_t, \dots, x_{t +\text{\textturntwo}}) := 0    
\end{equation}
where $\text{\textturntwo}$ is the masking window selected from a uniform distribution from $0$ to the maximum time mask parameter $\mathbb{TM}$. The time position $t$ is picked from another uniform distribution over $[0, T)$ such that the maximum sequence length $T$ is not exceeded (i.e. if $t +\text{\textturntwo} > T$, we set it to $T$).

\textbf{Frequency Masking}: Frequency masking is applied such that $\phi$ consecutive frequency channels $[f, f +\phi)$ are masked, where $\phi$ is selected from a uniform distribution from 0 to the frequency mask parameter $\mathbb{FM}$, and $f$ is chosen from $[0, \nu)$, where $\nu$ is the input feature dimension, e.g. the number of MFCC channels. For raw waveform as input, $\nu=1$. Similar to time masking, if $f +\phi > \nu$, we set it to $f=\nu$.

\onecolumn
\section{Automatic Speech Recognition}
\label{sec:ASR}
\subsection{Overview}
ASR is traditionally formulated within a statistical framework. Formally, let $x_1^{T}$ denote a sequence of acoustic feature vectors, where $x_t \in \mathbb{R}^D$ for $1 \leq t \leq T$, extracted from the raw speech waveform via a feature extraction process (e.g. MFCC). Let $V$ represent the vocabulary set. Typically, each vector $x_t$ encodes information corresponding to a fixed-duration frame of the speech signal, such as 10 milliseconds.

By Bayes' decision rule \cite{Bayes:1763}, given the observed acoustic feature sequence $x_1^{T}$, an ASR system aims to determine the most probable word sequence $\hat{w}_1^{\hat{N}} \in V^*$ such that:
\begin{align}
\hat{w}_1^{\hat{N}} &= \arg\max_{N, w_1^N} p(w_1^N \mid x_1^T) \label{end-to-end-asr}  \\
&= \arg\max_{N, w_1^N} \left[ \frac{p(x_1^T \mid w_1^N) \cdot p(w_1^N)}{p(x_1^T)} \right] \\
&= \arg\max_{N, w_1^N} \left[ \frac{p(x_1^T \mid w_1^N) \cdot p(w_1^N)}{\text{const}(w_1^N)} \right]  \\
&= \arg\max_{N, w_1^N} \left[ p(x_1^T \mid w_1^N) \cdot p(w_1^N) \right] \label{classical-asr} 
\end{align}

where $p(w_1^N \mid x_1^{T})$ denotes the posterior probability of the word sequence $w_1^N$ of $N$ length conditioned on the acoustic features $x_1^{T}$.

The effectiveness of an ASR system is typically quantified using the Word Error Rate (WER), defined for a reference word sequence $\tilde{w}_1^{\tilde{N}}$ and a hypothesis $w_1^N$ produced by the system as:
\begin{equation}
    \text{WER} = \frac{S_w + D_w + I_w}{\tilde{N_w}},
\end{equation}
where $S_w$, $D_w$, and $I_w$ represent the minimal number of \textit{substitution}, \textit{deletion}, and \textit{insertion} operations, respectively, required to transform the reference sequence into the hypothesis. The quantity $S_w + D_w + I_w$ corresponds to the Levenshtein distance \cite{Levenshtein1965BinaryCC} between the two sequences. For an evaluation corpus containing multiple references, the numerator and denominator are computed by summing over all hypotheses and references, respectively. WER is typically reported as a percentage.

Conventional ASR architectures, as discussed in~\cite{ney1990acoustic}, employ the decision rule in Eq.~\ref{classical-asr}, wherein the acoustic likelihood $p(x_1^{T} \mid w_1^N)$ (the \textit{acoustic model}) and the prior $p(w_1^N)$ (the \textit{language model}) are modeled independently. In this context, the acoustic model is instantiated by wav2vec 2.0, while the language model is often implemented using count-based methods \cite{KneserNey95}.

\subsection{Language Modeling}
\label{sec:intro-lm}
We consider the task of language modeling due to its close relationship with ASR. A \textit{language model} (LM) defines a probability distribution over a label sequence $w_1^N$, denoted as $p_{LM}(w_1^N)$. This probability is typically factorized in an autoregressive fashion, although alternative non-autoregressive modeling approaches have also been proposed \cite{irie18_interspeech, devlin2019bert}:
\begin{equation}
    p_{LM}(w_1^N) = \prod_{n=1}^{N} p_{LM}(w_n | w_0^{n-1}), \label{lm-autoregressive}
\end{equation}
where the LM estimates the conditional probability $p_{LM}(w_n | w_1^{n-1})$. Traditional LMs rely on count-based methods under the $k$-th order Markov assumption, i.e., $p_{LM}(w_n | w_1^{n-1}) \approx p_{LM}(w_n | w_{n-k}^{n-1})$. In contrast, contemporary neural LMs are designed to leverage the full left context to directly model $p_{LM}(w_n | w_1^{n-1})$. To ensure that the normalization condition $\sum_{w_1^N} p_{LM}(w_1^N) = 1$ holds, all sequences are required to terminate with a special end-of-sequence (EOS) symbol.

The performance of an LM is commonly assessed via its perplexity (PPL) \cite{Jelinek1977PerplexityaMO}, which for a sequence $w_1^N$ is defined as:
\begin{equation}
    \text{PPL} = \left[ \prod_{n=1}^{N} p_{LM}(w_n | w_1^{n-1}) \right]^{-\frac{1}{N}} = \exp\left( -\frac{1}{N} \sum_{n=1}^N \log p_{LM}(w_n | w_1^{n-1}) \right).
\end{equation}
This formulation generalizes to a corpus-level evaluation by averaging the negative log probabilities of all tokens (along with their left contexts) across the corpus. Perplexity can be interpreted as the average effective number of choices the LM considers when predicting the next token. Lower perplexity indicates a better-performing model.

In Hidden Markov Model (HMM)-based ASR systems, the LM is an integral component. Although sequence-to-sequence (seq2seq) models do not incorporate an LM explicitly, empirical results have demonstrated that incorporating an external LM during decoding can significantly reduce the WER \cite{hwang2017character, hori2017advances, kannan2018analysis}, assuming no domain mismatch. Consequently, it is now standard practice to integrate an external LM into the decoding process of seq2seq ASR models, which is also the approach adopted in this thesis. In wav2vec 2.0 experiments, researchers usually consider three types of LMs: a count-based Kneser-Ney smoothed $n$-gram model \cite{KneserNey95}, an LSTM-based LM \cite{sundermeyer2012lstm}, and a Transformer-based LM \cite{irie2019language}.

\onecolumn
\section{Connectionist Temporal Classification (CTC)}
\label{sec:intro-ctc}
wav2vec 2.0 uses CTC to model, thus we provide an overview of CTC in this section.

\subsection{Topology}
A CTC model \cite{graves2012connectionist} consists of an encoder network followed by a linear projection and a softmax activation layer. The encoder takes as input a sequence of acoustic feature vectors $x_1^T$ and produces a corresponding sequence of hidden representations $h_1^{T'}$:

\begin{equation}
    h_1^{T'} = \text{Encoder}(x_1^T)
\end{equation}

where each encoding vector $h_t \in \mathbb{R}^{D_{\text{enc}}}$ for $1 \leq t \leq T'$, and $D_{\text{enc}}$ denotes the dimensionality of the encoder output. The length $T'$ of the output sequence is typically less than or equal to $T$, due to potential downsampling mechanisms, i.e., $T' \leq T$, and generally $T' < T$.

Let $V$ denote the vocabulary of permissible labels, and let $\varepsilon$ represent a special label not included in $V$. Define the extended label set as $V' = V \cup \{ \varepsilon \}$, where $\varepsilon$ is referred to as the \textit{blank} label, typically interpreted as representing either silence or the absence of a label. The output of the encoder network is processed through a linear transformation followed by a softmax activation, yielding:

\begin{equation}
    o_1^{T'} = \text{Softmax}(\text{Linear}(h_1^{T'}))
\end{equation}

where $o_t \in [0, 1]^{|V'|}$ for $1 \leq t \leq T'$. The $k$-th component of the output vector $o_t$, denoted $o_{t,k}$, corresponds to the probability of emitting the $k$-th label from $V'$ at time step $t$:

\begin{equation}
    o_{t,k} = p_t(v_k \mid h_1^{T'}),
\end{equation}

with $v_k \in V'$ and $1 \leq k \leq |V'|$. This formulation characterizes the output distribution of a CTC model, specifying a per-frame categorical distribution over the extended label set $V'$, including the blank label.

Given this frame-level distribution, the CTC model defines a probability distribution over all possible output label sequences $w_1^N$ conditioned on the input $x_1^T$, formally expressed as $p_{\text{CTC}}(w_1^N \mid x_1^T) := p_{\text{CTC}}(w_1^N \mid h_1^{T'})$. To construct this distribution, define a \textit{path} as a label sequence $y_1^{T'}$ of length $T'$ such that each $y_t \in V'$ corresponds to a label emitted at time step $t$.

Under the CTC framework, a key assumption is that of \textit{conditional independence} across time steps, implying that the joint probability of a path $y_1^{T'}$ conditioned on the encoder outputs factorizes as follows:

\begin{equation}
    p(y_1^{T'} \mid h_1^{T'}) = \prod_{t=1}^{T'} p_t(y_t \mid h_1^{T'}).
\end{equation}

A path $y_1^{T'}$ can be formally regarded as an \textit{alignment} corresponding to an output label sequence. Specifically, let $\mathcal{B} : (V')^* \rightarrow V^*$ denote the \textit{collapse} function, which operates by first merging consecutive repeated labels and subsequently removing all blank symbols. For instance, consider the examples:
\[
\mathcal{B}(\varepsilon \varepsilon c c \varepsilon a \varepsilon a a a \varepsilon t t t t \varepsilon) = \mathcal{B}(c c c \varepsilon a \varepsilon a a a t t) = caat.
\]
Under this definition, any path $y_1^{T'}$ satisfying $\mathcal{B}(y_1^{T'}) = w_1^N$ serves as a valid alignment for the label sequence $w_1^N$. The probability assigned to a label sequence $w_1^N$ is obtained by marginalizing over all its possible alignments:

\begin{equation}
\begin{split}
p_{\text{CTC}}(w_1^N \mid h_1^{T'}) &= \sum_{y_1^{T'}: \mathcal{B}(y_1^{T'}) = w_1^N} p(y_1^{T'} \mid h_1^{T'}) \\
&= \sum_{y_1^{T'}: \mathcal{B}(y_1^{T'}) = w_1^N} \prod_{t=1}^{T'} p_t(y_t \mid h_1^{T'})    
\end{split}
\label{ctc-def}
\end{equation}

CTC loss for the input-target pair \((x_1^{T}, w_1^N)\) is defined as the negative log-likelihood of the target sequence under the CTC model, i.e., the cross-entropy loss: $-\log p_{\text{CTC}}(w_1^N \mid x_1^{T})$.

\begin{figure}[h]
    \centering
    \includegraphics[width=\linewidth]{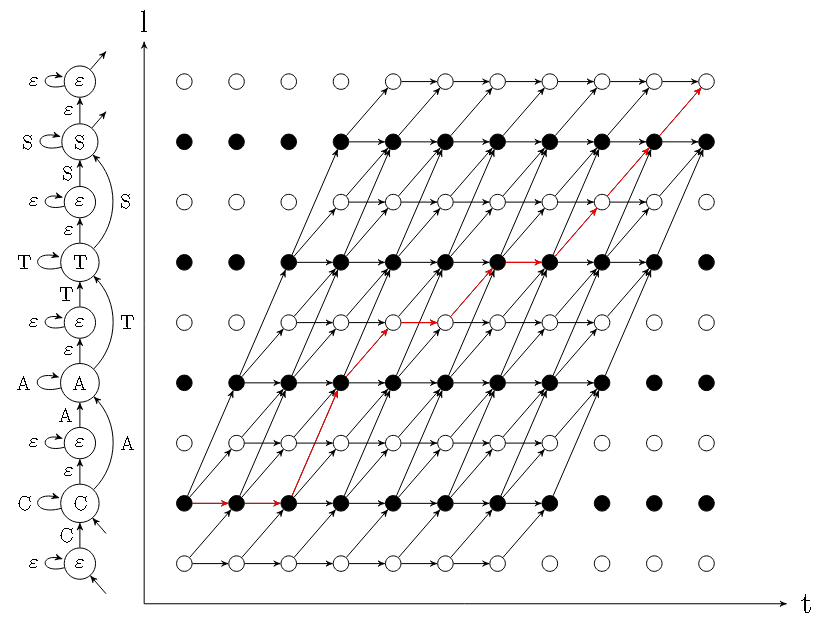}
    \caption{An illustrative CTC lattice, adapted from \cite{graves2006connectionist}, depicts all valid alignment paths corresponding to the label sequence $CATS$ along the $l$-axis, distributed over 11 time frames on the $t$-axis. Each node $(t, l)$ denotes the emission of label $l$ at time step $t$. Black nodes correspond to the emission of true (non-blank) labels, whereas white nodes indicate the emission of the blank symbol. The highlighted red path exemplifies a possible alignment: $CCCA\varepsilon\varepsilon TT\varepsilon S\varepsilon$. On the left, a finite state machine is shown, which constrains valid transitions through the lattice, subject to appropriate initial and terminal states.}
    \label{fig:ctc-lattice}
\end{figure}

An illustrative example of the CTC topology is depicted in Figure~\ref{fig:ctc-lattice}. As shown, the corresponding lattice structure admits two valid initial nodes and two valid final nodes. This arises from the fact that a valid alignment path may begin or end with either a true label or the special blank label, reflecting the inherent flexibility of CTC in handling variable-length alignments.

We highlight two properties of CTC that ensure its consistency with the ASR task:
\begin{itemize}
    \item The CTC alignment $y_1^{T'}$, as previously defined, is \textit{strictly monotonic}.
    \item The conditional probability $p_{\text{CTC}}(w_1^N \mid h_1^{T'})$ defines a distribution over all label sequences with $N \leq T'$, aligning with typical ASR scenarios where $N < T'$.
\end{itemize}
Additionally, CTC exhibits an empirically observed ``peaky'' behavior \cite{graves2006connectionist}, wherein it predominantly emits the blank symbol with high probability, interspersed with sharp peaks corresponding to predicted labels. This behavior diverges from the intuitive expectation that a label should be strongly emitted throughout its spoken duration. A formal analysis of this phenomenon is provided in \cite{zeyer2021:peakyctc}.

\subsection{CTC Forward-Backward Algorithm}
The training objective of a CTC model is to minimize the negative log-likelihood $-\log p_{\text{CTC}}(w_1^N \mid h_1^{T'})$, which necessitates the computation of $p_{\text{CTC}}(w_1^N \mid h_1^{T'})$. A direct evaluation using the definition in Equation \ref{ctc-def} is computationally intensive due to the exponential number of possible alignments $y_1^{T'}$ corresponding to the target sequence $w_1^N$. To address this, \citet{graves2006connectionist} proposed an efficient dynamic programming (DP) algorithm, analogous to the forward-backward procedure employed in HMMs \citep{rabiner1989tutorial}, to compute this quantity.

For a given label sequence $w_1^N$, we define the forward variables $Q_{\varepsilon}(t, n)$ and $Q_{l}(t, n)$ for all $1 \leq t \leq T'$ and $0 \leq n \leq N$ as the total probability of all valid alignments of the partial sequence $w_1^n$ from frame $1$ to frame $t$, where the alignment ends at frame $t$ with either a blank symbol ($\varepsilon$) or a non-blank label ($l$), respectively. Formally:
\begin{equation}
    Q_{\varepsilon}(t, n) = \sum_{\substack{y_1^t, \mathcal{B}(y_1^t) = w_1^n \\ y_t = \varepsilon}} \prod _{t'=1}^t p_{t'}(y_{t'} | h_1^{T'})
\end{equation}
\begin{equation}
    Q_{l}(t, n) = \sum_{\substack{y_1^t, \mathcal{B}(y_1^t) = w_1^n \\ y_t \neq \varepsilon}} \prod _{t'=1}^t p_{t'}(y_{t'} | h_1^{T'})
\end{equation}

here, $w_1^0$ denotes the empty sequence. The DP procedure is initialized using the following base cases:

\begin{align}
    &Q_{\varepsilon}(t, 0) = \prod_{t'=1}^t p_{t'}(\varepsilon | h_1^{T'})  \quad \forall 1 \leq t \leq T' \\
    &Q_{l}(t, 0) = 0  \quad \forall 1 \leq t \leq T' \\
    &Q_{\varepsilon}(1, n) = 0 \quad \forall 1 \leq n \leq N \\
    &Q_{l}(1, n) = \begin{cases}
        p_1(w_1 | h_1^{T'}), & \text{if } n = 1 \\
        0, & \text{if } 2 \leq n \leq N
    \end{cases} \label{ql1n-init}
\end{align}

For all \( t \geq 2 \) and \( n \geq 1 \), the values \( Q_{\varepsilon}(t, n) \) and \( Q_{l}(t, n) \) can be computed using the following DP recursion:

\begin{align} 
    Q_{\varepsilon}(t, n) &= p_t(\varepsilon | h_1^{T'}) \cdot [Q_{\varepsilon}(t-1, n) + Q_{l}(t-1, n)] \\
    Q_{l}(t, n) &= p_t(w_n | h_1^{T'}) \cdot \left[Q_{l}(t-1, n) + Q_{\varepsilon}(t-1, n-1) 
    + \overline{Q}_{l}(t-1, n-1)
    \right],
\end{align}

where $\overline{Q}_{l}(t-1, n-1)$ is defined as:
\begin{equation}
    \overline{Q}_{l}(t-1, n-1) = \begin{cases}
    Q_{l}(t-1, n-1), & \text{if } w_n \neq w_{n-1} \\
    0, & \text{otherwise}
    \end{cases}
\end{equation}

By the definition of the forward variables, $p_{\text{CTC}}(w_1^N | h_1^{T'})$ could be calculated as follows:
\begin{equation}
    p_{\text{CTC}}(w_1^N | h_1^{T'}) = Q_{\varepsilon}(T', N) + Q_{l}(T', N)
\end{equation}

Similarly, the backward variables $R_{\varepsilon}(t, n)$ and $R_{l}(t, n)$, defined for all $1 \leq t \leq T'$, $1 \leq n \leq N+1$, represent the total alignment probabilities corresponding to the decoding of the label sequence $w_n^N$ from frame $t$ to frame $T'$, conditioned on the assumption that the label emitted at frame $t$ is either a blank symbol ($\varepsilon$) or a true label ($l$), respectively.

\begin{equation}
    R_{\varepsilon}(t, n) = \sum_{\substack{y_1^t, \mathcal{B}(y_1^t) = w_n^N \\ y_t = \varepsilon}} \prod _{t'=t}^{T'} p_{t'}(y_{t'} | h_1^{T'})
\end{equation}

\begin{equation}
    R_{l}(t, n) = \sum_{\substack{y_1^t, \mathcal{B}(y_1^t) = w_n^N \\ y_t \neq \varepsilon}} \prod _{t'=t}^{T'} p_{t'}(y_{t'} | h_1^{T'})
\end{equation}
where $w_{N+1}^N$ is seen as the empty sequence. The following initializations are needed for the DP:
\begin{align}
    &R_{\varepsilon}(t, N+1) = \prod_{t'=t}^{T'} p_{t'}(\varepsilon | h_1^{T'})  \quad \forall 1 \leq t \leq T' \\
    &R_{l}(t, N+1) = 0  \quad \forall 1 \leq t \leq T' \\
    &R_{\varepsilon}(T', n) = 0 \quad \forall 1 \leq n \leq N \\
    &R_{l}(T', n) = \begin{cases}
        p_{T'}(w_N | h_1^{T'}), & \text{if } n = N \\
        0, & \text{if } 1 \leq n \leq N-1
    \end{cases}\label{rltn-init}
\end{align}

For $t \leq T'-1, n \leq N $, $R_{\varepsilon}(t, n)$ and $R_{l}(t, n)$ could be later computed by the DP recursion as follows:
\begin{align} 
    R_{\varepsilon}(t, n) &= p_t(\varepsilon | h_1^{T'}) \cdot [R_{\varepsilon}(t+1, n) + R_{l}(t+1, n)] \\
    R_{l}(t, n) &= p_t(w_n | h_1^{T'}) \cdot \left[R_{l}(t+1, n) + R_{\varepsilon}(t+1, n+1) 
    + \overline{R}_{l}(t+1,n+1)
    \right],
\end{align}
where $\overline{R}_{l}(t+1, n+1)$ is defined as:
\begin{equation}
\overline{R}_{l}(t+1,n+1) =  \begin{cases}
    R_{l}(t+1, n+1), & \text{if } w_{n} \neq w_{n+1} \\
    0, & \text{otherwise}
    \end{cases}
\end{equation}
By the definition of the backward variables, $p_{\text{CTC}}(w_1^N | h_1^{T'})$ could be calculated as:
\begin{equation}
    p_{\text{CTC}}(w_1^N | h_1^{T'}) = R_{\varepsilon}(1, 1) + R_{l}(1, 1)
\end{equation}
Besides calculating $p_{\text{CTC}}(w_1^N | h_1^{T'})$, \citet{graves2006connectionist} shows that these forward and backward variables could be employed to compute the error signal of the model during training analytically.

\subsection{CTC Decoding}
Decoding (also referred to as \textit{search}, \textit{recognition}, or \textit{inference}) denotes the process of identifying the word sequence $\hat{w}_1^{\hat{N}}$ that satisfies the decision rule defined in Equation \ref{end-to-end-asr}, given a sequence of acoustic feature vectors $x_1^{T}$. In this section, we first present general forms of decision rules applicable to seq2seq models. Subsequently, we detail specific decision rules and corresponding search algorithms tailored to the CTC framework.

\subsubsection{Decision Rules}
\label{search:decision-rule}

Because seq2seq models directly calculate $p(w_1^N | x_1^{T})$, the decision rule for the search is defined as belows:
\begin{equation}
    \hat{w}_1^{\hat{N}} = \arg\max_{N, w_1^N} p(w_1^N | x_1^{T}) = \arg\max_{N, w_1^N} \left[\log p(w_1^N | x_1^{T})\right]. \label{seq2seq-decision-rule}
\end{equation}

As for CTC, decision rule \ref{seq2seq-decision-rule} could be rewritten as:
\begin{equation}
    \hat{w}_1^{\hat{N}} = \arg\max_{N, w_1^N} \left[ \sum_{y_1^{T'}: \mathcal{B}(y_1^{T'}) = w_1^N} \prod_{t=1}^{T'} p_t(y_t | h_1^{T'}) \right]. \label{ctc-decision-rule}
\end{equation}

Because of the peaky behavior \cite{zeyer2021:peakyctc}, CTC models usually generate blanks more and true labels less. In order to minimize blanks and increase true label emission, an alternative \textit{prior correction} could be employed \cite{manohar15_interspeech}. The prior of a label $y$, which could then be either a true label or the blank label, is calculated as:
\begin{equation}
    p_\text{prior}(y) = \frac{1}{\sum_{\tilde{h}_1^{\tilde{T'}}}\tilde{T'}}    \sum_{\tilde{h}_1^{\tilde{T'}}} \sum_{t=1}^{\tilde{T'}} p_t(y | \tilde{h}_1^{\tilde{T'}}),
\end{equation}
where the encoded vectors $\tilde{h}_1^{\tilde{T}}$ are retrieved based on the empirical distribution. The label posteriors in decision rule \ref{ctc-decision-rule} of the search are then divided by their priors as belows:
\begin{equation}
    \hat{w}_1^{\hat{N}} = \arg\max_{N, w_1^N} \left[ \sum_{y_1^{T'}: \mathcal{B}(y_1^{T'}) = w_1^N} \prod_{t=1}^{T'} \frac{p_t(y_t | h_1^{T'})}{p_\text{prior}^{\lambda_{\text{prior}}}(y_t)}  \right],
\end{equation}
where $\lambda_{\text{prior}}$ is the prior scale, which is also a constant. In previous empirical studies, prior correction has been used in order to boost CTC performance.

As detailed in Section \ref{sec:ASR} above, combining an external LM $p_{\text{LM}}$ with a seq2seq model generally leads to enhanced results. The most simple and commonly employed method is \textit{shallow fusion} \cite{gulcehre2015using}, which simply linearly combines decision rule in Equation \ref{seq2seq-decision-rule} with the LM score as shown below:
\begin{equation}
    \hat{w}_1^{\hat{N}} = \arg\max_{N, w_1^N} \left[\log p(w_1^N | x_1^{T}) + \lambda_{\text{LM}} \log p_{\text{LM}}(w_1^N)\right], \label{shallow-fusion}
\end{equation}
where $\lambda_{\text{LM}}$ is the LM scale. It is important to be aware that there should be no domain mismatch between the LM and the inference domain, and the external LM should preserve the linguistic distribution in the inference domain well. For CTC, shallow fusion could be simply combined with prior correction.

\subsubsection{Greedy Search}
Greedy search is a simplified decoding strategy for CTC, in which the summation over all alignment paths in the decision rule in Equation \ref{ctc-decision-rule} is approximated by the most probable path alone. That is, it selects the single highest-probability alignment path predicted by the CTC model:

\begin{equation}
\begin{split}
    \hat{w}_1^{\hat{N}} &\approx \arg\max_{N, w_1^N} \left[ \max_{y_1^{T'}: \mathcal{B}(y_1^{T'}) = w_1^N} \prod_{t=1}^{T'} p_t(y_t | h_1^{T'}) \right] \\
    &= \mathcal{B} \left(\arg\max_{y_1^{T'}} \left[ \prod_{t=1}^{T'} p_t(y_t | h_1^{T'}) \right]\right).
\end{split}
\end{equation}

This can be computed in a straightforward manner by selecting the maximum label emission at each time frame, followed by the application of the collapse function to the resulting alignment. Greedy search offers the advantage of high computational efficiency, particularly when employed with Transformer-based architectures, as the encoder computations can be parallelized across time frames. Nevertheless, greedy search typically yields lower accuracy compared to beam search methods \cite{graves2006connectionist}, and it lacks the capability to incorporate an external LM.

\subsubsection{Time-synchronous Search}
CTC is inherently a time-synchronous model, making it natural to adopt a time-synchronous decoding strategy. As proposed in \cite{GravesJaitly14}, a time-synchronous beam search variant for CTC is introduced, which estimates the sum of \textit{some} path probabilities corresponding to each hypothesis by maintaining two auxiliary partial probabilities during DP. Most studies adopt a variant of beam search in which the score assigned to each hypothesis corresponds to the probability of its most likely alignment path, augmented with an external LM score. This approach bears resemblance to Viterbi decoding employed in HMM-based systems \cite{rabiner1989tutorial}. Incorporating prior correction, the resulting decision rule for this decoding procedure can be expressed as:

\begin{equation}
    \hat{w}_1^{\hat{N}} = \arg\max_{N, w_1^N} \left[\log \left( \max_{y_1^{T'}: \mathcal{B}(y_1^{T'}) = w_1^N} \prod_{t=1}^{T'} \frac{p_t(y_t | h_1^{T'})}{p_{\text{prior}}^{\lambda_{\text{prior}}}(y_t)} \right) + \lambda_{\text{LM}} \log p_{\text{LM}}(w_1^N) \right].
\end{equation}

Details of the search procedure are provided in Algorithm \ref{ctc-time-sync-algo}. The initial hypothesis consists of the empty sequence augmented with a virtual begin-of-sequence (BOS) label. It is important to note that, in this context, the set of partial hypotheses corresponds not to partial label sequences, but to partial CTC alignments. 

At each iteration over the input frames, the scores of all possible hypothesis expansions are computed. Subsequently, if multiple expansions yield the same label sequence, only the one with the highest score is retained. This operation constitutes a form of maximum recombination over all expansions that correspond to the same label sequence. Following this recombination, a subset of the expansions is selected based on the beam size constraint, thereby pruning the search space. 

In the absence of an external LM and prior correction, this procedure reduces to the standard greedy search.

\begin{algorithm}[h]
\caption{CTC time-synchronous beam search.}
\label{ctc-time-sync-algo}
\begin{algorithmic}[1]
\State \textbf{Input:} beam size $H$, CTC posterior $p_t(v | h_1^{T'})\forall v \in V', 1\leq t \leq T'$, language model $p_{\text{LM}}$, LM scale $\lambda_{\text{LM}}$, label prior $p_{\text{prior}}(v)\forall v \in V'$, prior scale $\lambda_{\text{prior}}$
\State \textbf{Initialize:} \\
$hyps \leftarrow \{\texttt{<BOS>}\}$ \Comment{set of partial hypotheses} \\
$scores \leftarrow \{\}$ \Comment{hash-map storing scores of partial hypotheses} \\
$scores[\texttt{<BOS>}] \leftarrow 0$
\For{$t = 1 \dots T'$}
\\
    \texttt{/* Consider all possible hypothesis expansions */}
    \State $expansions \leftarrow \{ g\cdot v | g \in hyps, v \in V'  \}$
    \For{$g \in hyps$}
        \For{$v \in V'$}     
            \State $scores[g\cdot v] \leftarrow scores[g] + \log p_t(v | h_1^{T'}) -\lambda_{\text{prior}} \log p_{\text{prior}}(v)$
            \If {$v \neq \varepsilon$ and $v \neq \text{last label of }g$} 
                \State $scores[g\cdot v] \leftarrow scores[g\cdot v] + \lambda_{\text{LM}} \log p_{\text{LM}}(v | \mathcal{B}(g))$
            \EndIf
        \EndFor
    \EndFor
\\
\texttt{/* Recombination step */}
    \For{$g_1, g_2 \in expansions \times expansions$}
        \If{$g_1 \neq g_2$ and $\mathcal{B}(g_2) = \mathcal{B}(g_1)$}
            \If{$scores[g_2] > scores[g_1]$}
                \State $scores[g_1] \leftarrow -\infty$
            \Else
                \State $scores[g_2] \leftarrow -\infty$
            \EndIf
        \EndIf
    \EndFor
\\
\texttt{/* Pruning step */}
    \State $hyps \leftarrow $ the $H$ partial hypotheses in $expansions$ with highest $scores$
\EndFor
\State \textbf{Return:} $\mathcal{B}\left( \arg\max_{y_1^{T'} \in hyps} scores[y_1^{T'}] \right)$
\end{algorithmic}
\end{algorithm}

\subsubsection{Label-synchronous Search}
Label-synchronous search was originally introduced alongside the CTC topology in \cite{graves2006connectionist}. This approach directly applies the decision rule given in Equation \ref{seq2seq-decision-rule}, wherein beam search operates at the level of label sequences. To facilitate this procedure, the \textit{CTC prefix probability} is required, denoted by
\begin{equation}
    p_{\text{CTC}}(w_1^n, * \mid x_1^{T}) := p_{\text{CTC}}(w_1^n, * \mid h_1^{T'})
\end{equation}
which quantifies the probability that the CTC model emits a sequence beginning with the prefix \( w_1^n \), followed by an arbitrary suffix (including the empty suffix).

\begin{equation}
    p_{\text{CTC}}(w_1^n, * | h_1^{T'}) = \sum_{\nu \in (V')^{*}}p_{\text{CTC}}(w_1^n\cdot \nu | h_1^{T'}), \label{ctc-prefix-prob}
\end{equation}

where $(V')^{*}$ denotes the set of all label sequences over  $V'$, including the empty sequence. The score assigned to a partial hypothesis $w_1^n$ is defined as $\log p_{\text{CTC}}(w_1^n, * | h_1^{T'})$. If the hypothesis $w_1^n$ is complete, its final score becomes $\log p_{\text{CTC}}(w_1^n | h_1^{T'})$. The label-synchronous beam search procedure, along with the DP formulation for computing the CTC prefix probability during decoding, is detailed in \cite{graves2012connectionist}.

The label-synchronous search procedure operates at the output label level, facilitating straightforward integration of an external LM. The partial score of a hypothesis $w_1^n$ is given by:
\begin{equation}
    \log p_{\text{CTC}}(w_1^n, * | h_1^{T'}) + \lambda_{\text{LM}}\log p_{\text{LM}}(w_1^n).
\end{equation}

This approach sums over all path probabilities corresponding to a partial label sequence, but incurs a time complexity of $\mathcal{O}(T')$ per hypothesis extension. Consequently, the total complexity becomes $\mathcal{O}(N_{\text{max}} T')$, where $N_{\text{max}}$ denotes the length of the longest hypothesis. This is substantially less efficient than the $\mathcal{O}(T')$ complexity of time-synchronous search. Moreover, CTC is inherently time-synchronous.


\newpage
{
    \bibliographystyle{ieeenat_fullname}
    \bibliography{IEEEFull}
}
\end{document}